\documentclass[11pt]{article}

\usepackage[preprint]{acl}
\usepackage{paralist}
\usepackage{times}
\usepackage{latexsym}
\usepackage{booktabs}
\usepackage[table]{xcolor}
\usepackage{amssymb}
\usepackage{amsmath}
\usepackage{caption}
\usepackage{tabularx}
\usepackage{multirow}
\usepackage[table]{xcolor}
\usepackage[T1]{fontenc}

\usepackage[utf8]{inputenc}

\usepackage{microtype}

\usepackage{inconsolata}

\usepackage{graphicx}

\newcommand{\systemname}{\textbf{VizDAGuide}}

\title{VizDAGuide: Misleading Chart Question Answering in Compact Vision Language Models via DAG-Structured Supervision}

\author{
  \textbf{Haiyang Zhang}$^{1}$\thanks{Equal contribution.}, 
  \textbf{Sicheng Song}$^{1}$\footnotemark[1]\thanks{Corresponding author.}, 
  \textbf{Zixin Chen}$^{2}$, 
  \textbf{Jie Gu}$^{1}$, \\
  \textbf{Huiya Ou}$^{1}$, 
  \textbf{Chenhui Li}$^{1}$, 
  \textbf{Changbo Wang}$^{1}$ \\
  $^{1}$East China Normal University \\
  $^{2}$The Hong Kong University of Science and Technology \\
  \texttt{\{Oceanz, jiegu, 10245501484\}@stu.ecnu.edu.cn} \\
  \texttt{scsong@dase.ecnu.edu.cn}, \texttt{zchendf@connect.ust.hk} \\
  \texttt{chli@cs.ecnu.edu.cn}, \texttt{cbwang@cs.ecnu.edu.cn}
}

\begin{document}
\maketitle
\begin{abstract}
Misleading charts can lead vision-language models to conclusions that conflict with underlying data.
Existing work focuses on inference-time strategies, while training-time supervision remains less explored.
We present \systemname{}, a DAG-structured supervision method for misleading-chart question answering.
The method represents intermediate reasoning states and their dependencies as a directed acyclic graph, assigns supervision weights according to their roles, and serializes the graph into a unified training target for open-ended and multiple-choice QA with single-pass inference.
Experiments on two benchmarks across Qwen3-VL and InternVL3.5 models show improvements over a content- and weight-matched Linear CoT-SFT baseline on both answer formats.
Process-level evaluation shows gains in grounding and operation execution, while ablations indicate that dependency structure and weighted supervision are most effective when combined.
These results support DAG-structured weighted supervision for misleading-chart question answering.
\end{abstract}

\section{Introduction}
\label{sec:introduction}
\begin{figure}[t]
    \centering
    \includegraphics[width=\columnwidth]{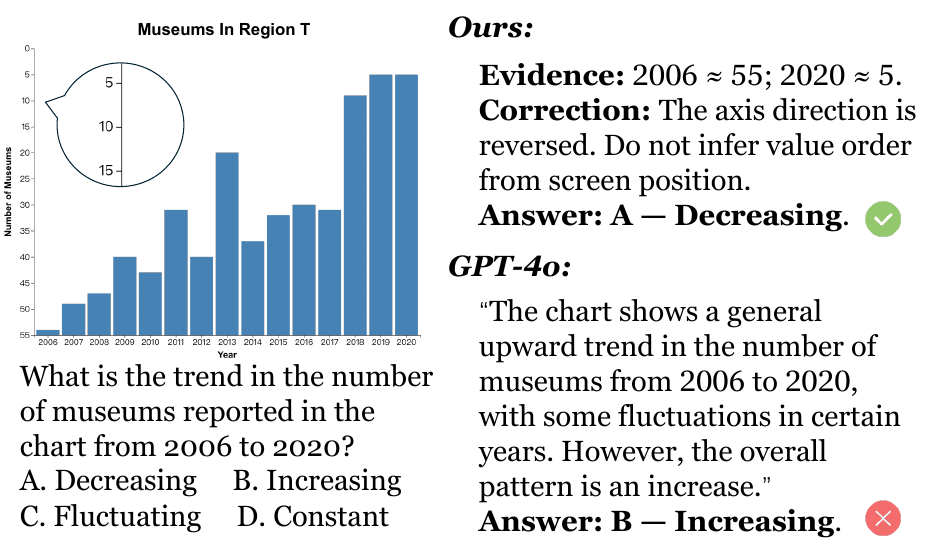}
    \caption{An example multiple-choice question (MCQ)
illustrating misleading-chart reasoning.
The reversed axis makes the chart visually suggest
an increasing trend despite decreasing underlying
values.}
    \label{fig:example}
\end{figure}

Charts support data analysis and communication by
encoding numerical information through visual marks.
However, misleading designs can create a mismatch
between visual appearance and the underlying data.
Figure~\ref{fig:example} illustrates this
problem: although the bars visually suggest an
increasing trend, the reversed axis indicates that
the underlying values decrease.
A model that follows the apparent visual pattern can
therefore reach the opposite conclusion from one that
grounds the values and accounts for the misleading
encoding.
More generally, inverted axes, inconsistent scales,
and other misleading designs can distort the inference
suggested by a chart.
For multimodal language models, answering questions
about such charts requires both extracting relevant
information and determining whether the visual
encoding supports the intended inference.
Recent benchmarks show that misleading charts remain
challenging for multimodal models in chart question
answering
\citep{chen-etal-2025-unmasking,
ortizbarajas-etal-2026-chartattack,
mahbub2025}.

Existing datasets present this problem through different answer formats, including open-ended questions and multiple-choice questions.
These formats determine how a model expresses its answer, but they do not define all the abilities needed to obtain it.
For example, computing a difference and selecting the option that describes that difference both require identifying the relevant entities, reading their values, and applying the correct operation.
When the chart is misleading, both tasks also require recognizing the visual risk and selecting an appropriate correction.
This motivates a shared representation of misleading chart reasoning beneath the downstream answer format.

Intermediate supervision provides a way to expose these abilities during training.
Prior work transfers chart reasoning through synthesized rationales \citep{carbune-etal-2024-chart}, while ChartR decomposes questions to assess the correctness of intermediate steps \citep{chen2026chartr}.
Building on this perspective, we organize misleading chart reasoning around question scoping, evidence grounding, visual-risk diagnosis, correction selection, operation execution, decision formation, and answer generation, and supervise how these abilities contribute to answers across both question formats.

A further challenge concerns the relationships among these abilities.
Their dependencies involve branching and convergence rather than only a sequence of adjacent steps.
Question scope determines both which evidence to select and which operation to perform.
The operation then combines the selected evidence with the requirements of the question.
Visual-risk diagnosis provides a separate basis for deciding whether correction is needed and which visual inference should be rejected.
For example, comparing two series on a dual-axis chart requires combining the question scope with values read from the corresponding axes; comparing their screen heights alone is insufficient.
A textual rationale can describe this process, but its sentence order need not make these dependencies explicit. This issue is relevant for compact multimodal models, whose reasoning performance can be sensitive to the structure and complexity of supervision.
Prior work shows that smaller models may struggle to learn from long or undifferentiated reasoning traces, motivating more explicit and selective reasoning supervision \citep{li-etal-2025-small-models,zhang-etal-2024-tinychart,chen-etal-2025-skip}.

We propose \systemname{}, a weighted DAG-structured supervision method for misleading chart question answering. 
Each node represents an intermediate reasoning state associated with an ability, and typed relations specify how these states depend on one another. 
We serialize the graph into text and use it as the training target for a multimodal language model. 
The same structured representation is applied across both open-ended and multiple-choice chart question answering. The graph therefore provides an explicit supervision structure while remaining compatible with autoregressive generation. We evaluate both final-answer accuracy and intermediate reasoning performance, and use controlled comparisons to assess the effect of graph-structured supervision. Our contributions are summarized as follows:
\begin{compactitem}
\item We formulate misleading chart question answering as a set of intermediate abilities and construct a unified supervision scheme for both open-ended and multiple-choice chart question answering.
\item We propose \systemname{}, a weighted DAG-structured supervision method that models dependencies among intermediate reasoning states and enables structured reasoning to be learned through autoregressive training. We further validate its effectiveness across model scales and VLM families.
\item We introduce a process-level evaluation scheme for misleading chart question answering that complements answer-level accuracy by assessing the quality of intermediate reasoning states, enabling a more comprehensive evaluation of model reasoning behavior.
\end{compactitem}

\section{Related Work}
\label{sec:related_work}

\subsection{Chart Question Answering}

Chart question answering requires models to locate chart elements, retrieve values, and perform numerical or logical reasoning.
Early benchmarks use synthetic charts and template-based questions \citep{kafle2018dvqa,methani2020plotqa}, while later work extends chart QA to human-written questions, structured chart representations, and chart-specific reasoning objectives \citep{masry-etal-2022-chartqa,singh-shekhar-2020-stl,liu-etal-2023-deplot,carbune-etal-2024-chart,liu-etal-2023-matcha,zhang-etal-2024-tinychart}.
Our work considers the additional challenge posed by visual encodings that can bias the interpretation of chart data~\citep{bharti2025chartomvisualtheoryofmindbenchmark}.

\subsection{Misleading Chart Question Answering}

Misleading charts can cause models to rely on visual cues that conflict with the underlying data.
Misleading ChartQA evaluates this problem with multiple-choice questions across chart and misleader types \citep{chen-etal-2025-unmasking}.
ChartAttack studies the generation and impact of misleading charts and introduces AttackViz for evaluating the errors induced by different misleaders \citep{ortizbarajas-etal-2026-chartattack}.
Other work explores inference-time mitigation, including chart-to-table conversion and chart reconstruction \citep{tonglet-etal-2026-protecting,tonglet-etal-2026-chart,zhang2026navigatingmiragedualpathagentic}.
These studies provide benchmarks and mitigation strategies for misleading chart reasoning, while our work focuses on structuring the reasoning supervision used during training~\citep{zhang2026vizanchordecodingmanipulationintent}.

\subsection{Structured Reasoning Supervision}

Chain-of-thought represents reasoning as a sequence of intermediate steps \citep{wei2022chain}, leaving dependencies among these steps implicit.
Graph-based methods organize intermediate states with branching and aggregation, but mainly structure reasoning at inference time \citep{besta2024graph}.
UniCoTT further studies structured chain-of-thought distillation by transferring chain-, tree-, and graph-structured reasoning from large language models to smaller models with structural learning objectives \citep{zhuang2025unicott}.
For chart question answering, ChartR decomposes chart questions into sequential subquestions and shows that final-answer accuracy alone cannot reveal errors in intermediate reasoning \citep{chen2026chartr}.
Our work instead defines explicit dependencies among chart reasoning abilities and uses them as structured supervision for misleading chart question answering across answer formats.

\section{Method}
\subsection{Problem Formulation and Overview}

We consider misleading chart question answering with a chart image $I$, a question $q$, and an answer $a$.
For multiple-choice QA, we denote the candidate set as $C$; for open-ended QA, $C=\varnothing$.
Both formats are thus represented by
\begin{equation}
    x=(I,q,C).
\end{equation}

Rather than supervising only the final answer, our method associates each instance with intermediate reasoning states and their dependencies.
Given a shared ability set $\mathcal{A}$, we denote the instantiated reasoning states for $x$ as
\begin{equation}
    S(x)=\{s_k \mid A_k\in\mathcal{A}\}.
\end{equation}
\begin{figure*}[t]
    \centering
    \includegraphics[width=0.95\textwidth]{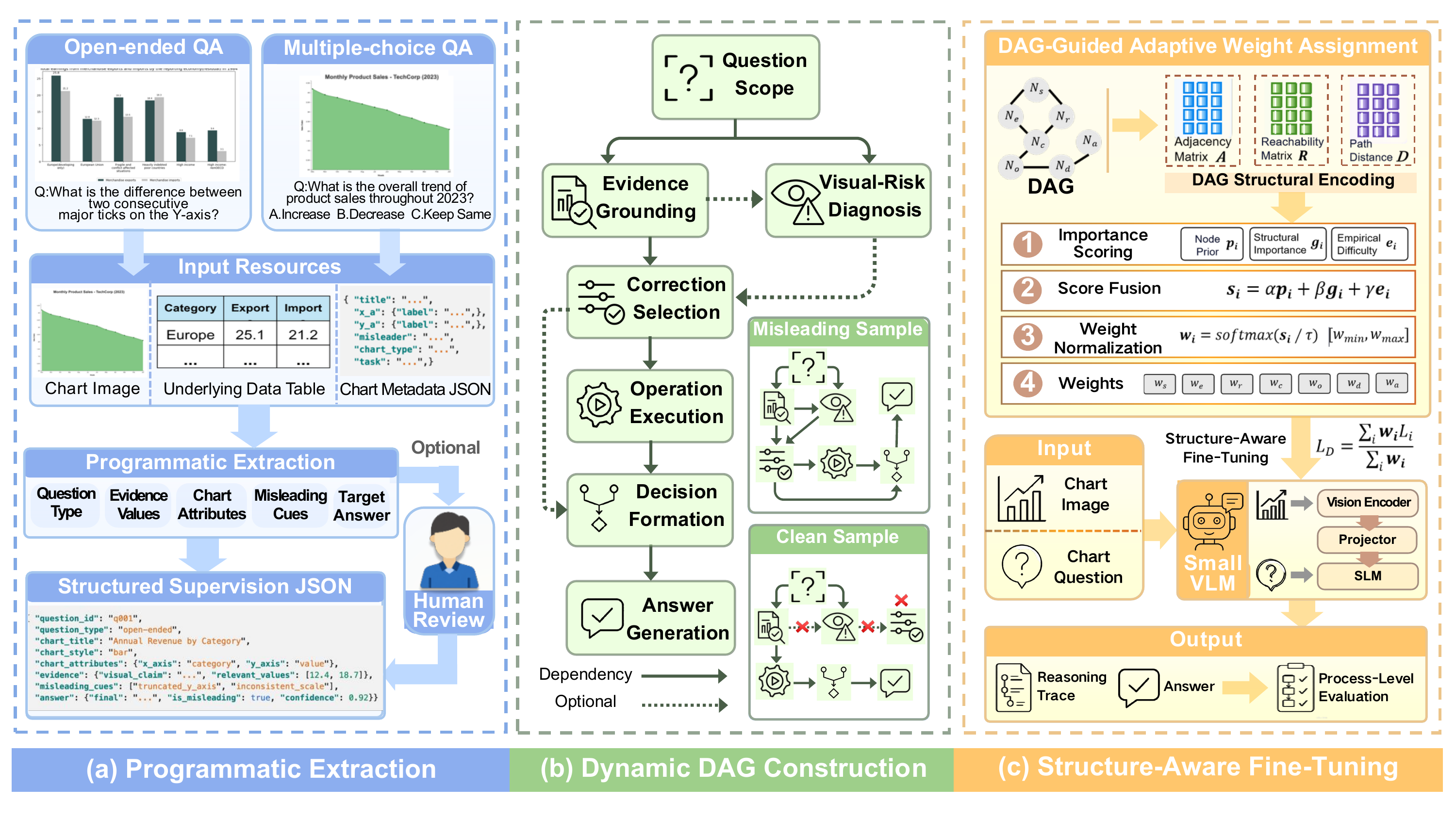}
    \caption{Overview of VizDAGuide.}
    \label{fig:overview}
\end{figure*}
As shown in Figure~\ref{fig:overview}, we first instantiate these abilities from the training data, organize their dependencies as a DAG, and serialize the graph as an autoregressive training target.
The resulting supervision is shared across open-ended and multiple-choice QA, while the final answer follows the format of each dataset.
\subsection{Shared Reasoning Abilities}
\label{abilities}

We define the ability set to satisfy three requirements: it should cover the reasoning process from question interpretation to answer generation, distinguish different reasoning steps, and apply to both answer formats.
The decomposition follows the functional demands of the task.
Chart QA requires interpreting the question, grounding relevant chart evidence, and executing numerical or logical operations \citep{masry-etal-2022-chartqa}.
Misleading charts additionally require identifying visual cues that can distort the required inference and, when necessary, selecting an appropriate correction \citep{chen-etal-2025-unmasking}.
We further separate decision formation from answer generation, allowing the same reasoning state to support both open-ended and multiple-choice outputs.
This results in seven shared abilities,
\begin{equation}
\begin{aligned}
\mathcal{A} = \{&
A_{\mathrm{scope}}, A_{\mathrm{evidence}}, A_{\mathrm{risk}},
A_{\mathrm{correction}},\\
&
A_{\mathrm{operation}}, A_{\mathrm{decision}},
A_{\mathrm{answer}}\}.
\end{aligned}
\end{equation}

Question Scoping identifies task requirements, Evidence Grounding
grounds relevant evidence, Visual-Risk Diagnosis detects misleading
cues, Correction Selection determines the required correction,
Operation Execution performs the required operation, Decision Formation
derives the conclusion, and Answer Generation produces the final answer.
For each ability $A_k$, its instantiated state is written as $s_k=(A_k,c_k)$, where $c_k$ denotes the supervision content for the current instance.
The decomposition serves as a task-oriented supervision schema rather than an exhaustive taxonomy of chart reasoning.

These abilities capture distinct reasoning functions, but their dependencies are not strictly linear.
For example, evidence grounding is conditioned on the question scope, while operation execution may depend on both the selected evidence and a correction for misleading visual cues.
Correction is also conditional: it is required only when the visual encoding can affect the inference needed by the question.
A fixed reasoning chain does not make these dependencies explicit.

We therefore represent the instantiated reasoning states as a directed acyclic graph
\begin{equation}
    G_x=(V_x,E_x),
\end{equation}
where $V_x\subseteq S(x)$ contains the reasoning states used for instance $x$, and $E_x$ specifies their directed dependencies.
An edge $(s_i,s_j)\in E_x$ indicates that the content of $s_i$ provides supervision required for constructing or interpreting $s_j$.
The graph remains acyclic because dependencies follow the progression from question interpretation and evidence grounding to decision and answer generation.

The graph structure is instance-dependent.
For a clean chart, reasoning may proceed from grounded evidence to the required operation without activating a correction state.
For a misleading chart, visual-risk diagnosis can activate a correction path that constrains the subsequent operation or decision.
We also annotate dependencies that should not support a downstream state when a misleading visual cue would otherwise provide a shortcut.
Figure~\ref{fig:overview}(b) illustrates these structures.
\subsection{Structured Target Construction}

We construct the structured supervision target from the resources provided by each dataset.
Depending on availability, these resources include the chart image, underlying data table, chart metadata, the question and answer, and annotations of misleading cues.
A programmatic extraction stage maps these resources to the fields required by the shared ability schema, including question type, relevant evidence, chart attributes, misleading cues, and the target answer.
The extracted fields are checked for consistency before being used to instantiate the reasoning states.

For each instance, the resulting node contents and dependencies form a target graph $G_x$.
We serialize each DAG into a textual target for autoregressive training.
\begin{equation}
    y_x=\operatorname{Serialize}(G_x,a),
\end{equation}
where $a$ is the dataset answer. The serialization contains node and relation spans that preserve both reasoning content and dependency structure.
\subsection{VizDAGuide: Weighted DAG Supervision}
\label{Weighted DAG Supervision}
We assign different supervision weights to reasoning states according to their roles in the DAG.
For each target graph, we derive its adjacency matrix $A$, reachability matrix $R$, and path-distance matrix $D$, from which the structural importance of each node is estimated.
The score of node $i$ combines its ability prior $p_i$, structural importance $g_i$, and empirical difficulty $e_i$:
\begin{equation}
    z_i=\alpha p_i+\beta g_i+\gamma e_i .
\end{equation}

We normalize the scores and map them to a bounded weight range:
\begin{equation}
\begin{aligned}
    \tilde{w}_i
    &= \frac{\exp(z_i/\tau)}
    {\sum_j \exp(z_j/\tau)}, \\
    w_i
    &= w_{\min}
    +(w_{\max}-w_{\min})\tilde{w}_i .
\end{aligned}
\end{equation}
where $\tau$ controls the weight distribution.
The node weights are assigned to their corresponding token spans, with separate weights used for relation and answer spans.

Given target tokens $y_{1:T}$, we optimize
\begin{equation}
    \mathcal{L}_{\mathrm{DAG}}
    =
    \frac{\sum_{t=1}^{T} w_t
    \mathcal{L}_{\mathrm{CE}}(y_t,\hat{y}_t)}
    {\sum_{t=1}^{T} w_t},
\end{equation}
where $w_t$ is the weight of the span containing token $t$.
\paragraph{Training and inference.}
Given input $x$, the model autoregressively predicts the structured target $y_x$ and is optimized with $\mathcal{L}_{\mathrm{DAG}}$.
At inference time, only $x$ is provided, and the trained model generates the reasoning trace and final answer in a single pass.

\section{Experiments}

\subsection{Experimental Setup}

\paragraph{Datasets.}
We evaluate on both open-ended and multiple-choice misleading chart QA.
For open-ended QA (OEQ), we use the AttackViz/PlotQA data \citep{ortizbarajas-etal-2026-chartattack}, with 4,000 training instances and the full official test split of 6,042 questions.
Among them, 5,961 have valid reference answers and are used for answer-level evaluation.
For multiple-choice QA (MCQ), we use the benchmark introduced by \citet{chen-etal-2025-unmasking}.
Its 3,055 questions are split by chart group and misleader type into 2,138 training and 917 test instances; 880 test questions have scorable answers.
We ensure no train--test overlap in question IDs, chart images, or underlying tables.

\paragraph{Models and training.}
Our main experiments use Qwen3-VL-4B-Instruct and Qwen3-VL-8B-Instruct, fine-tuned with LoRA.We additionally evaluate InternVL3.5-4B-Instruct as an answer-level cross-backbone validation.

\paragraph{Baselines.}
We compare against the base model and answer-only supervised fine-tuning. For intermediate reasoning supervision, we fine-tune a Linear CoT-SFT baseline following prior work~\citep{ho-etal-2023-large}. It preserves the same intermediate content and applies the same weights to corresponding spans, while removing node identifiers and explicit dependency relations, providing a content- and weight-matched control for DAG structure. We further compare unweighted and weighted DAG variants to isolate the effect of supervision weighting within the same structure.
For MCQ, we also compare with the VLM SFT setting reported in ChartCynics~\citep{zhang2026navigatingmiragedualpathagentic}.
As its resources are not released, we use the reported result on the 305-sample test set as a reference.

\paragraph{Evaluation metrics.}
For open-ended QA, we report strict final-answer accuracy (FAA) and relaxed accuracy with a 5\% relative tolerance for numerical answers (RA).
For multiple-choice QA, we report answer accuracy and misleading-choice rate (MCR).
For intermediate reasoning performance, we evaluate Grounding, Operation Type, Correction, and Process Accuracy.
We report Clean False Correction (Clean FC) for clean charts; for MCQ, we report Sufficiency, which measures whether the available evidence supports the final decision.
\paragraph{External-model process evaluation.}
External models are not required to follow our DAG serialization.
Instead, we use a unified structured protocol that elicits evidence, operation, visual correction, decision, and final answer in a model-agnostic format.
These outputs are then evaluated with the same process-level criteria used for our models.
Additional training details, metric definitions, and external-model evaluation protocols are provided in the Appendix.
\subsection{Main Results}

\noindent\textbf{(1) Intermediate reasoning supervision.} Table~\ref{tab:ablation} compares supervision strategies on open-ended and multiple-choice QA.
Answer-only supervision yields a small FAA gain, whereas Linear CoT-SFT substantially improves both FAA and RA.
This indicates that intermediate reasoning supervision, rather than answer supervision alone, accounts for much of the improvement.

\noindent\textbf{(2) Contribution of DAG structure and weighted supervision.}
With matched intermediate content and span weights, DAG-structured supervision outperforms Linear CoT-SFT across both tasks, model scales, and backbones.
For the 4B model, weighted supervision also outperforms the unweighted DAG, showing complementary gains from dependency structure and weighting.
The same trend holds for InternVL3.5-4B.
On MCQ, the Qwen3-VL results are numerically higher than the 68.52\% SFT result reported by ChartCynics~\citep{zhang2026navigatingmiragedualpathagentic}.

\noindent\textbf{(3) Accuracy and misleading-choice rate.}
MCR shows a different pattern from answer accuracy.
For the 4B model, answer-only supervision has the lowest MCR (4.77\%) but only 72.05\% accuracy, with 23.18\% of predictions falling into other wrong options.
Linear CoT-SFT and the DAG variant have similar MCR values, while the latter achieves higher accuracy (74.55\% vs.\ 71.93\%).
The 8B model shows the same pattern, with higher accuracy despite a higher MCR for the DAG variant.
Thus, MCR reflects selection of the annotated misleading option rather than overall question accuracy.
\begin{table}[t]
    \centering
    \footnotesize
    \setlength{\tabcolsep}{3.2pt}
    \renewcommand{\arraystretch}{1.08}
    \caption{Comparison of supervision variants across VLM backbones.}
    \label{tab:ablation}
    
    \begin{tabular}{@{}lcccc@{}}
        \toprule
        & \multicolumn{2}{c}{\textbf{Open-ended QA}}
        & \multicolumn{2}{c}{\textbf{MCQ}} \\
        \cmidrule(lr){2-3}
        \cmidrule(l){4-5}
        \textbf{Setting}
        & \textbf{FAA} $\uparrow$
        & \textbf{RA} $\uparrow$
        & \textbf{Acc}. $\uparrow$
        & \textbf{MCR} $\downarrow$ \\
        \midrule

        \rowcolor{gray!10}
        \multicolumn{5}{l}{\textit{Qwen3-VL-4B}} \\

        Base
        & 34.98 & 45.03 & 62.73 & 17.61 \\

        + Answer-only-SFT
        & 36.02 & 47.07 & 72.05 & \textbf{4.77} \\

        + Linear CoT-SFT
        & 56.25 & 68.28 & 71.93 & 15.45 \\

        + U-DAG
        & 55.31 & 67.89 & 70.68 & 15.91 \\

        \textbf{+ VizDAGuide (ours)}
        & \textbf{57.54}
        & \textbf{69.97}
        & \textbf{74.55}
        & 15.34 \\

        \midrule

        \rowcolor{gray!10}
        \multicolumn{5}{l}{\textit{Qwen3-VL-8B}} \\

        Base
        & 41.94 & 52.56 & 54.09 & 29.09 \\

        + Linear CoT-SFT
        & 57.10 & 69.47 & 71.93 & \textbf{15.68} \\

        \textbf{+ VizDAGuide (ours)}
        & \textbf{59.54}
        & \textbf{71.88}
        & \textbf{74.20}
        & 16.93 \\

        ChartCynics SFT
        & -- & -- & 68.52 & -- \\

        \midrule
        
        \rowcolor{gray!10}
        \multicolumn{5}{l}{\textit{InternVL3.5-4B}} \\

        Base
        & 40.92 & 50.18 & 52.61 & 28.18 \\

        + Linear CoT-SFT
        & 52.83 & 64.55 & 70.80 & \textbf{14.20} \\

        \textbf{+ VizDAGuide (ours)}
        & \textbf{53.97}
        & \textbf{65.64}
        & \textbf{72.95}
        & 15.80 \\

        \bottomrule
    \end{tabular}
\end{table}

\subsection{Process-level Analysis}

\begin{table}[htbp]
\centering
\small
\setlength{\tabcolsep}{2pt}
\renewcommand{\arraystretch}{1.03}

\caption{Process-level results on open-ended and multiple-choice QA.
Gnd., Op., Corr., Pac., FC, and Suff. denote Grounding, Operation,
Correction, Process Accuracy, Clean FC, and Sufficiency.}
\label{tab:process_results}

\begin{tabular*}{\columnwidth}
{@{\extracolsep{\fill}}lccccc@{}}
\toprule
\multicolumn{6}{c}{\textbf{Open-ended QA}} \\
\midrule
\textbf{Supervision}
& \textbf{Gnd.}$\uparrow$
& \textbf{Op.}$\uparrow$
& \textbf{Corr.}$\uparrow$
& \textbf{Pac.}$\uparrow$
& \textbf{FC}$\downarrow$ \\
\midrule

\rowcolor{gray!10}
\multicolumn{6}{l}{\textit{Qwen3-VL-4B}} \\

Linear CoT-SFT
& 57.49 & 68.21 & \textbf{88.64} & 48.99 & 11.10 \\

U-DAG
& 57.57 & 67.99 & 86.66 & 49.72 & 2.67 \\

\textbf{VizDAGuide (Ours)}
& \textbf{59.44} & \textbf{70.07} & 84.90
& \textbf{50.53} & \textbf{1.81} \\

\midrule

\rowcolor{gray!10}
\multicolumn{6}{l}{\textit{Qwen3-VL-8B}} \\

Linear CoT-SFT
& 59.54 & 69.45 & 83.69 & 48.97 & 8.90 \\

\textbf{VizDAGuide (Ours)}
& \textbf{61.38} & \textbf{71.97} & \textbf{89.92}
& \textbf{54.08} & \textbf{2.90} \\

\bottomrule
\end{tabular*}

\vspace{4pt}

\begin{tabular*}{\columnwidth}
{@{\extracolsep{\fill}}lccccc@{}}
\toprule
\multicolumn{6}{c}{\textbf{Multiple-choice QA}} \\
\midrule
\textbf{Supervision}
& \textbf{Gnd.}$\uparrow$
& \textbf{Op.}$\uparrow$
& \textbf{Corr.}$\uparrow$
& \textbf{Suff.}$\uparrow$
& \textbf{Pac.}$\uparrow$ \\
\midrule

\rowcolor{gray!10}
\multicolumn{6}{l}{\textit{Qwen3-VL-4B}} \\

Linear CoT-SFT
& 70.45 & 33.75 & 91.36 & 94.66 & 19.09 \\

U-DAG
& 66.82 & 38.64 & 82.16 & 94.09 & 20.23 \\

\textbf{VizDAGuide (Ours)}
& \textbf{78.64} & \textbf{54.55} & \textbf{95.45}
& \textbf{96.70} & \textbf{41.59} \\

\midrule

\rowcolor{gray!10}
\multicolumn{6}{l}{\textit{Qwen3-VL-8B}} \\

Linear CoT-SFT
& 69.55 & 29.66 & 89.43 & 94.77 & 15.68 \\

\textbf{VizDAGuide (Ours)}
& \textbf{76.82} & \textbf{53.64} & \textbf{92.05}
& \textbf{97.16} & \textbf{38.64} \\

\bottomrule
\end{tabular*}

\end{table}

\label{Process_analysis}
\noindent\textbf{(1) Grounding and operation.}
Table~\ref{tab:process_results} shows that the main process-level changes occur in Grounding and Operation.
On open-ended QA, DAG-structured supervision improves both metrics over Linear CoT-SFT at both model scales.
The gains are larger on MCQ, where Operation rises from 33.75\% to 54.55\% for 4B and from 29.66\% to 53.64\% for 8B.
Grounding also improves at both scales, suggesting gains in evidence selection and operation execution.

\noindent\textbf{(2) Correction behavior.}
Correction remains high across supervision settings.
For the 4B model on open-ended QA, Correction changes from 88.64\% to 84.90\%, while Clean FC drops from 11.10\% to 1.81\%.
Despite the lower Correction score, FAA and RA still improve over Linear CoT-SFT, indicating fewer unnecessary corrections on clean charts without loss in answer-level performance.
For the 8B model, Correction increases from 83.69\% to 89.92\% and Clean FC decreases from 8.90\% to 2.90\%, improving both correction accuracy and false-correction control.

\noindent\textbf{(3) Process accuracy across tasks.}
Process Accuracy changes across the two answer formats.
On OEQ, it increases from 48.99\% to 50.53\% for 4B, while on MCQ it rises from 19.09\% to 41.59\% for 4B and from 15.68\% to 38.64\% for 8B.
Correction and Sufficiency remain high, with smaller changes than Grounding and Operation.
The MCQ gains are associated with improved grounding and operation execution, while correction varies with chart condition and model scale.

\paragraph{Process--answer relation.}
Table~\ref{tab:process_answer_relation} shows different patterns of answer failure across tasks.
For OEQ, Operation is the main bottleneck at both model scales, with the highest correlation with answer correctness and a 100\% answer error rate when it alone fails.
Grounding and Correction failures are more recoverable, indicating the importance of operation execution after evidence grounding.

On MCQ, Sufficiency and Grounding show the strongest associations with answer correctness, while isolated Operation failures are less damaging.
Table~\ref{tab:misleader_process} shows a similar pattern across misleader types: OEQ is dominated by Operation, whereas MCQ is associated with Sufficiency and Correction.
This suggests that OEQ relies more on operation execution, while MCQ relies more on evidence sufficiency for the final decision.

\begin{table}[t]
    \centering
    \setlength{\tabcolsep}{4pt}
    \renewcommand{\arraystretch}{1.00}
    \caption{Process--answer relation. Corr. measures process--answer correctness correlation; Answer Error is the final-answer error rate when only the corresponding process is incorrect.}
    \label{tab:process_answer_relation}
    \small
    \begin{tabular}{@{}llcccc@{}}
        \toprule
        & & \multicolumn{2}{c}{\textbf{Corr.}}
        & \multicolumn{2}{c}{\textbf{Answer Error}} \\
        \cmidrule(lr){3-4}
        \cmidrule(l){5-6}
        \textbf{Task} & \textbf{Process}
        & \textbf{4B} & \textbf{8B}
        & \textbf{4B} & \textbf{8B} \\
        \midrule

        \multirow{3}{*}{OEQ}
        & Grounding
        & 0.447 & 0.439
        & 9.94 & 8.68 \\

        & Operation
        & \textbf{0.757} & \textbf{0.756}
        & \textbf{100.00} & \textbf{100.00} \\

        & Correction
        & 0.105 & 0.085
        & 18.44 & 18.88 \\

        \midrule

        \multirow{4}{*}{MCQ}
        & Grounding
        & 0.294 & 0.212
        & 57.14 & 48.57 \\

        & Operation
        & 0.174 & 0.129
        & 20.66 & 22.13 \\

        & Correction
        & 0.123 & 0.067
        & 62.50 & 36.36 \\

        & Sufficiency
        & \textbf{0.316} & \textbf{0.290}
        & \textbf{100.00} & \textbf{100.00} \\

        \bottomrule
    \end{tabular}
\end{table}
\begin{table}[htbp]
\centering
\small
\setlength{\tabcolsep}{2pt}
\renewcommand{\arraystretch}{1.03}

\caption{Top-5 misleader types ranked by average process--answer correlation
across 4B and 8B. Process denotes the highest-correlated process.}
\label{tab:misleader_process}

\begin{tabular*}{\columnwidth}
{@{\extracolsep{\fill}}cllcc@{}}
\toprule
\textbf{Task}
& \textbf{Misleader Type}
& \textbf{Process}
& \textbf{4B}
& \textbf{8B} \\
\midrule

\multirow{5}{*}{OEQ}
& Stacked misuse
& Operation & \textbf{0.963} & \textbf{0.960} \\

& Axis-range misuse
& Operation & 0.937 & 0.946 \\

& Dual-axis misuse
& Operation & 0.930 & 0.910 \\

& 3D effect
& Operation & 0.919 & 0.909 \\

& Log-scale misuse
& Operation & 0.898 & 0.904 \\

\midrule

\multirow{5}{*}{MCQ}
& Small size
& Sufficiency & \textbf{1.000} & \textbf{1.000} \\

& Inappropriate aggregation
& Sufficiency & 0.697 & 0.697 \\

& Continuous encoding
& Correction & 0.807 & 0.477 \\

& Exceeding canvas
& Sufficiency & 0.461 & 0.608 \\

& Scale range
& Sufficiency & 0.389 & 0.316 \\

\bottomrule
\end{tabular*}

\end{table}

\subsection{Fine-grained Analysis}
\label{sec:fine_grained}

\begin{figure*}[t]
    \centering
    \includegraphics[width=\linewidth]{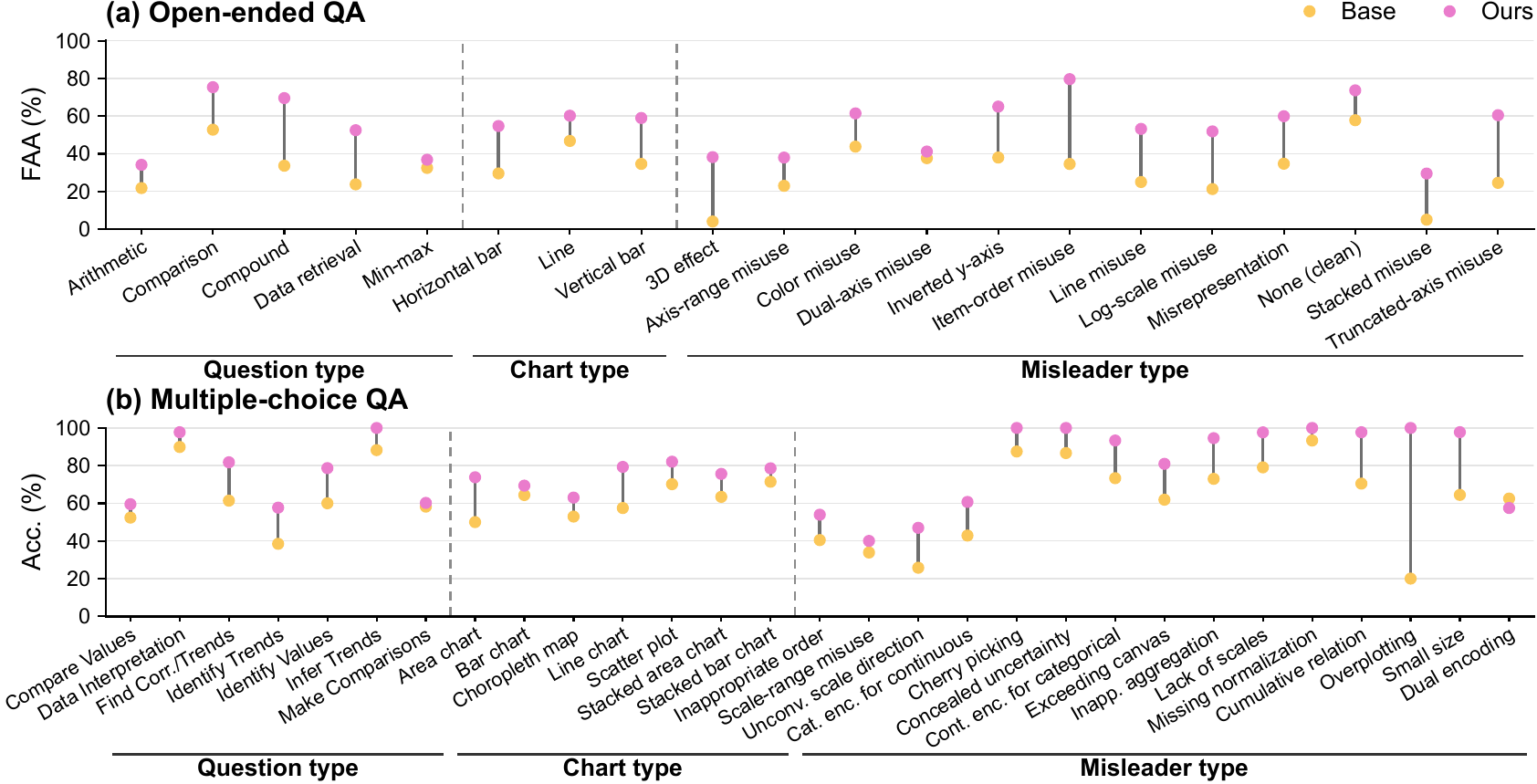}
    \caption{Fine-grained comparison between the Base model and VizDAGuide across question types, chart types, and misleading conditions.}
    \label{fig:comparison}
\end{figure*}

We further compare \systemname{} with the Qwen-VL-4B base model across question types, chart types, and misleading conditions in Figure~\ref{fig:comparison}

\paragraph{Question type.}
On open-ended QA, all five question types improve over Base.
The largest gains appear in compound and data-retrieval questions
(35.92\% and 28.77\%), while min--max remains the hardest category.
This suggests gains in both evidence retrieval and multi-step reasoning.

MCQ shows a different pattern.
Find Correlations/Trends gains the most, followed by Identify Values, while Make Comparisons changes less.
Find Correlations/Trends shows the clearest gain, which remains supported after Holm correction.
Thus, the improvements are spread across more question types in open-ended QA than in MCQ.
\paragraph{Chart type.}
On open-ended QA, all three chart types improve over Base.
Horizontal and vertical bars gain about 25 percentage points, while line charts gain 13.41\%; all three gains remain supported after Holm correction.

MCQ gains span area, line, bar, choropleth, scatter, and stacked charts.
Area and line charts show the largest gains and remain supported after correction.
Overall, the improvements extend across multiple chart representations rather than concentrating on one chart type.

\begin{figure}[t]
    \centering
    \includegraphics[width=\columnwidth]{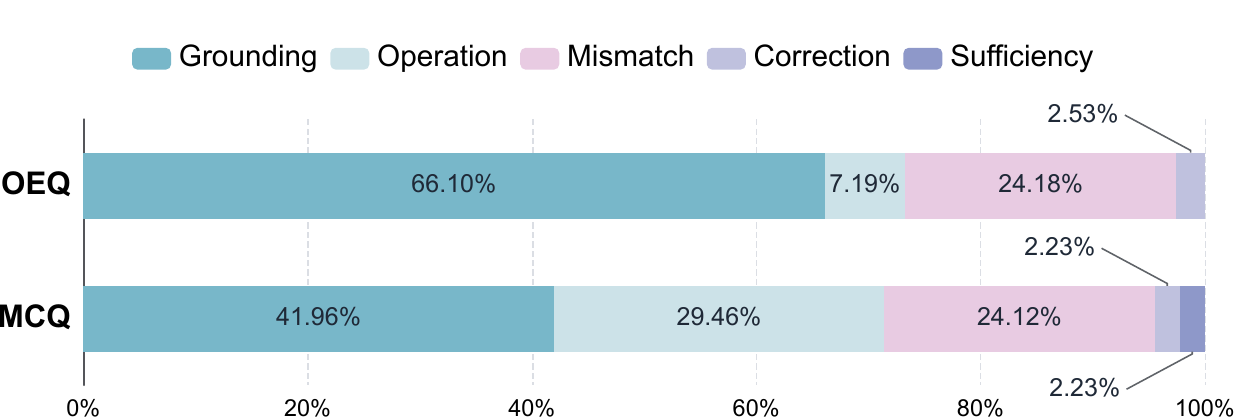}
    \caption{Distribution of the earliest identified error stage in VizDAGuide predictions on open-ended and multiple-choice QA.}
    \label{fig:error_analysis}
\end{figure}

\paragraph{Misleader type.}

\begin{table*}[t]
\centering
\scriptsize
\setlength{\tabcolsep}{2pt}
\renewcommand{\arraystretch}{1.00}

\begin{minipage}[t]{0.495\textwidth}
\vspace{0pt}
\centering

\begin{tabular*}{\linewidth}{@{\extracolsep{\fill}}lcccccc@{}}
\toprule
\multicolumn{7}{c}{\textbf{(a) Open-ended QA}} \\
\midrule
\textbf{Model}
& \textbf{RA} $\uparrow$
& \textbf{Gnd.} $\uparrow$
& \textbf{Op.} $\uparrow$
& \textbf{Corr.} $\uparrow$
& \textbf{Pac.} $\uparrow$
& \textbf{FC} $\downarrow$ \\
\midrule

GPT-5.5
& 64.42 & 59.34 & 55.36 & 63.16 & 36.54 & 4.62 \\

Gemini 3 Flash
& 64.28 & 59.45 & 55.86 & 61.48 & 37.12 & 3.76 \\

Kimi-K3
& 63.83 & 57.78 & 55.18 & 55.98 & 34.57 & 1.57 \\

Claude Opus 4.8
& 63.11 & 48.47 & 31.71 & 52.58 & 15.50 & 1.62 \\

Gemini 2.5 Pro
& 61.63 & 50.71 & 51.69 & 58.18 & 29.11 & 6.14 \\

o4-mini
& 61.08 & 44.66 & 44.37 & 36.60 & 21.89 & 0.33 \\

Gemini 2.5 Flash
& 59.52 & 52.58 & 51.03 & 30.25 & 18.27 & 1.00 \\

GPT-4.1
& 58.87 & 48.01 & 45.40 & 37.06 & 23.27 & \textbf{0.29} \\

Claude Sonnet 4.5
& 58.06 & 51.65 & 46.40 & 52.58 & 26.94 & 2.81 \\

Claude Haiku 4.5
& 54.45 & 38.90 & 31.91 & 41.67 & 15.94 & 1.33 \\

GPT-4o
& 53.75 & 38.77 & 39.17 & 36.74 & 19.23 & 0.71 \\

\midrule

Qwen3-VL-4B + ours
& 69.97 & 59.44 & 70.07 & 84.90 & 50.53 & 1.81 \\

Qwen3-VL-8B + ours
& \textbf{71.88}
& \textbf{61.38}
& \textbf{71.97}
& \textbf{89.92}
& \textbf{54.08}
& 2.90 \\

\bottomrule
\end{tabular*}

\end{minipage}
\hfill
\begin{minipage}[t]{0.495\textwidth}
\vspace{0pt}
\centering

\begin{tabular*}{\linewidth}{@{\extracolsep{\fill}}lcccccc@{}}
\toprule
\multicolumn{7}{c}{\textbf{(b) Multiple-choice QA}} \\
\midrule
\textbf{Model}
& \textbf{Acc.} $\uparrow$
& \textbf{Gnd.} $\uparrow$
& \textbf{Op.} $\uparrow$
& \textbf{Corr.} $\uparrow$
& \textbf{Suff.} $\uparrow$
& \textbf{Pac.} $\uparrow$ \\
\midrule

GPT-5.5
& \textbf{80.00}
& 50.57
& \textbf{73.07}
& 47.84
& 93.52
& 24.89 \\

Kimi-K3
& 77.39 & 50.68 & 67.16 & 43.18 & 91.70 & 18.41 \\

Claude Opus 4.8
& 76.48 & 53.64 & 65.00 & 47.39 & 91.14 & 17.61 \\

Gemini 2.5 Pro
& 75.68 & 48.18 & 62.73 & 41.93 & 87.05 & 18.30 \\

Gemini 3 Flash
& 75.45 & 51.93 & 66.36 & 42.50 & 87.39 & 20.23 \\

Claude Sonnet 4.5
& 74.66 & 52.16 & 62.27 & 33.86 & 92.05 & 18.86 \\

Gemini 2.5 Flash
& 68.18 & 39.66 & 56.25 & 26.02 & 84.77 & 10.11 \\

Claude Haiku 4.5
& 66.02 & 52.39 & 51.48 & 28.18 & 85.45 & 8.07 \\

o4-mini
& 65.00 & 21.82 & 40.68 & 9.66 & 84.20 & 2.84 \\

GPT-4.1
& 64.32 & 31.82 & 60.34 & 16.14 & 84.09 & 10.00 \\

GPT-4o
& 62.95 & 27.95 & 58.30 & 18.30 & 84.20 & 10.68 \\

\midrule

Qwen3-VL-4B + ours
& 74.55
& \textbf{78.64}
& 54.55
& \textbf{95.45}
& 96.70
& \textbf{41.59} \\

Qwen3-VL-8B + ours
& 74.20
& 76.82
& 53.64
& 92.05
& \textbf{97.16}
& 38.64 \\

\bottomrule
\end{tabular*}

\end{minipage}

\caption{
Reference comparison with general-purpose models on open-ended and
multiple-choice misleading chart QA.
External models use a unified JSON evaluation protocol, whereas our models
use their native structured-output setting; the results are intended for
contextual comparison rather than strict ranking.
}
\label{tab:external_results}

\end{table*}

On open-ended QA, most misleading conditions improve over Base.
The largest gains appear in item-order, truncated-axis, 3D-effect, and log-scale misuse.
Stacked misuse rises from 4.98\% to 29.49\% but remains difficult.
Clean-chart accuracy also increases from 57.81\% to 73.67\%, showing that the gains are not limited to misleading cases.

MCQ shows a more varied pattern across misleader types.
Gains appear in unconventional scale directions, inappropriate aggregation, cumulative relationships, and several encoding-related conditions.
Dual encoding shows a numerical drop from 62.50\% to 57.50\%, but the difference is not supported after Holm correction, indicating no reliable degradation for this category.

\subsection{Error Analysis}
\label{sec:error_analysis}

We locate the earliest failed reasoning stage among incorrect predictions.
As shown in Figure~\ref{fig:error_analysis}, open-ended QA is dominated by grounding failures (66.10\%), while operation errors are relatively rare (7.19\%), indicating that many failures occur before the required computation is executed.
MCQ shows a different pattern: grounding remains the largest source of error (41.96\%), but operation failures rise to 29.46\%, suggesting that errors more often extend into operation execution after evidence has been identified.
Final-answer mismatches account for about one quarter of errors in both settings, whereas correction-related failures are uncommon.
Thus, the main bottleneck lies in grounding for open-ended QA, but is shared between grounding and operation in MCQ.

\subsection{Comparison with External Models}

Table~\ref{tab:external_results} contextualizes our trained models against general-purpose models evaluated on the same underlying test sets.
We re-evaluate these models using our evaluation pipeline and report both answer- and process-level metrics.
On open-ended QA, the 4B and 8B models trained via \systemname{} are competitive with the general-purpose models at the answer level and show stronger performance on several process metrics, with the 8B model achieving the highest RA and strong process-level results.
On MCQ, the trained models remain below the strongest general-purpose models in answer accuracy, but the gap is relatively small despite their smaller model scale.

\section{Conclusion}

We present \systemname{}, a weighted DAG-structured
supervision method that models dependencies among
intermediate reasoning abilities for misleading chart QA.
Across Qwen3-VL and InternVL3.5 models, the method
outperforms the content- and weight-matched Linear CoT-SFT
baseline at the answer level.
Process-level evaluation on Qwen3-VL shows gains in
grounding, operation execution, and process accuracy,
while ablations show complementary contributions from
dependency structure and weighted supervision.

Error analysis identifies grounding as the main
bottleneck on OEQ, while MCQ failures involve both
grounding and operation execution.
These findings highlight evidence grounding and
operation execution as directions for further improvement.

\section*{Limitations}
\paragraph{Dependence on structured training resources.}
Constructing VizDAGuide supervision relies on resources provided by the training benchmarks, such as underlying chart data, metadata, and misleading annotations when available. These resources are used only to construct training targets and are not required at inference time. Applying the method to datasets without such resources may therefore require additional
extraction or annotation procedures.
\paragraph{Supervision construction quality.}
VizDAGuide derives its reasoning graphs and supervision signals from structured chart information during training.
Errors or ambiguities in these intermediate resources can propagate into the constructed DAG targets and their supervision weights. Future work could incorporate uncertainty-aware construction or verification mechanisms to make DAG supervision more robust to imperfect training resources.


\bibliography{custom}

\appendix

\section{Dataset and Evaluation Coverage}
\label{app:data}

\paragraph{Open-ended QA.}
For open-ended QA (OEQ), we use the
AttackViz/PlotQA data, with 4,000 training
instances and the full official test split of
6,042 questions.
Among them, 5,961 have valid reference answers
and are used for answer-level evaluation.
The remaining 81 contain incomplete or
ambiguous reference annotations in the original
data and therefore do not support deterministic
answer-level scoring under our evaluation
protocol.
The eligibility criterion is determined solely
from the reference annotations and is fixed
across all models.
Strict final-answer accuracy (FAA) and relaxed
accuracy (RA) are computed on the same 5,961
questions.

\paragraph{Multiple-choice QA.}
For multiple-choice QA (MCQ), we use the
benchmark introduced by \citep{chen-etal-2025-unmasking}.
Its 3,055 questions are split by chart group
and misleader type into 2,138 training and
917 test instances.
Among the test questions, 880 have scorable
answers and are used for answer-level
evaluation.
The remaining 37 contain incomplete or
ambiguous reference annotations in the original
data and therefore do not support deterministic
answer-level scoring under our evaluation
protocol.
Answer accuracy and misleading-choice rate
(MCR) are computed on the same fixed set of
880 questions.
The MCQ test set contains misleading charts
only, so Clean False Correction (Clean FC) is
not applicable.

\paragraph{Scoring coverage.}
Answer eligibility is determined from the
reference annotations and is independent of
model predictions.
An empty, truncated, or unparseable response
on an answer-eligible question is counted as
incorrect rather than removed from the
denominator.
For process-level evaluation, each metric is
computed only when the reference annotations
contain the information required by that metric
and the metric is applicable to the instance.
The resulting applicability mask is fixed and
shared across all models.

\paragraph{Train--test separation.}
For both tasks, the predefined train--test
partitions are used throughout our experiments.
We ensure no train--test overlap in question IDs,
chart images, or underlying tables.

\paragraph{Training and inference resources.}
We construct the structured supervision targets
from the resources provided by each training
benchmark.
Depending on availability, these resources
include the chart image, underlying data table,
chart metadata, question and answer, and
annotations of misleading cues.
Test-set annotations are used only for offline
evaluation.
At inference time, the model receives the chart
image and question, together with the candidate
options for MCQ.
No underlying data tables, chart metadata,
misleading annotations, or evaluation references
are provided as inference-time inputs.

\section{DAG Construction and Serialization}
\label{app:dag}

\paragraph{Node contents.}
Our supervision schema defines seven reasoning-state
types corresponding to the shared abilities introduced
in Section~\ref{abilities}:
N1 for Question Scoping,
N2 for Evidence Grounding,
N3 for Visual-Risk Diagnosis,
N4 for Correction Selection,
N5 for Operation Execution,
N6 for Decision Formation, and
N7 for Answer Generation.
An instance activates the states applicable to its
reasoning process.
Question Scoping identifies the entities and relation
required by the question, while Evidence Grounding
records the relevant chart information.
Operation Execution specifies how the grounded
evidence is used, and Decision Formation expresses
the resulting task-level conclusion.
Answer Generation then renders this conclusion in
the format required by the dataset, as a free-form
answer for OEQ or a candidate selection for MCQ.

\paragraph{Dependency construction.}
We construct directed relations according to the
functional dependencies among the reasoning states.
Question scope guides evidence selection and
specifies the required operation.
Grounded evidence supplies the information used by
the operation.
When a misleading visual cue affects the required
inference, Visual-Risk Diagnosis activates a
Correction Selection state, which constrains the
subsequent operation or decision.
Operation Execution supports Decision Formation,
which in turn supports Answer Generation.
These dependencies follow the progression from
question interpretation and evidence grounding to
decision and answer generation, forming an acyclic
graph.

\paragraph{Conditional correction.}
The graph structure is instance-dependent.
For clean instances, reasoning can proceed from
question scope and grounded evidence to the required
operation without activating a correction state.
For misleading instances, Visual-Risk Diagnosis
activates a correction path that specifies which
visual inference should not be used and how the
subsequent reasoning should instead rely on the
grounded chart information.
We additionally record blocked-relation annotations
for misleading visual cues that could otherwise
provide an invalid shortcut to a downstream state.
These annotations are represented separately from
the active dependency edges and are not treated as
valid evidence dependencies.

\paragraph{Textual representation.}
We serialize the instantiated node contents together
with their parent references and typed dependency
relations.
Node identifiers distinguish reasoning states,
parent references expose their active dependencies,
and relation spans encode the corresponding graph
structure.
The following schematic example illustrates a
count question involving a logarithmic axis.
The values are illustrative and are not taken from
a test instance.

\begin{quote}
\small\ttfamily
\raggedright
N1 Question Scoping:
count values above 5000\\
N2 Evidence Grounding <- N1:
A=10400; B=9000; C=1000\\
N3 Visual-Risk Diagnosis:
logarithmic y-axis; visual distance is not linear
in value\\
N4 Correction Selection <- N3:
read the numerical values rather than comparing
linear visual distances\\
N5 Operation Execution <- N1,N2,N4:
count(value > 5000) = 2\\
N6 Decision Formation <- N5:
two values exceed the threshold\\
N7 Answer Generation <- N6:
2\\
Edges:
N1->N2[selects\_evidence];
N1->N5[specifies\_operation];
N2->N5[feeds\_evidence];
N3->N4[triggers\_correction];
N4->N5[constrains\_operation];
N5->N6[derives\_decision];
N6->N7[generates\_answer]\\
Blocked relation:
linear\_visual\_distance -> N5,
blocked by N4
\end{quote}

\paragraph{Matched Linear CoT control.}
We construct the Linear CoT baseline from the same
intermediate reasoning contents.
The corresponding fields are presented as
Question Scoping, Evidence Grounding,
Visual-Risk Diagnosis, Correction Selection,
Operation Execution, Decision Formation, and
Answer Generation.
We remove node identifiers, parent references,
explicit dependency relations, and
blocked-relation notation while preserving the
semantic content of the intermediate states.
Corresponding semantic spans receive the same
supervision weights as in the DAG target.
This construction is used for both OEQ and MCQ,
providing a content- and weight-matched control for
evaluating the contribution of explicit dependency
representation.

\paragraph{Role of the graph.}
The DAG serves as a training-time supervision
representation rather than an external
inference-time execution module.
Its serialized node and relation spans form the
autoregressive prediction target and receive the
structured supervision described in Section~\ref{Weighted DAG Supervision}.
At inference time, the model receives only the task
input and generates the intermediate reasoning
states and final answer in a single autoregressive
pass; no external graph executor or additional
correction pass is used.

\section{Training Details}
\label{app:training}
\subsection{DAG-Guided Adaptive Weight Assignment}
\label{app:adaptive_weights}

\paragraph{Structural encoding.}
For each instantiated target graph
$G_x=(V_x,E_x)$, we derive its adjacency
matrix $A$, reachability matrix $R$, and
directed path-distance matrix $D$ from the
active dependency relations.
The adjacency matrix records direct
dependencies between reasoning states,
whereas the reachability and path-distance
matrices capture transitive structural
relations within the DAG.
Blocked relations represent rejected
dependencies and are therefore excluded from
the active graph structure used to construct
these matrices.

\paragraph{Structural importance.}
The graph representation is used to estimate
the structural importance of each reasoning
state.
We denote the structural importance of node
$i$ as $g_i$.
It reflects the role of the node in the
instance-specific dependency structure based
on the adjacency, reachability, and
path-distance information derived from the
DAG.

\paragraph{Node importance score.}
Following Section~\ref{Weighted DAG Supervision},
the score of node $i$ combines its ability
prior $p_i$, structural importance $g_i$,
and empirical difficulty $e_i$:
\begin{equation}
z_i = \alpha p_i + \beta g_i + \gamma e_i.
\end{equation}
The three terms respectively capture the
reasoning role associated with the node,
its structural role in the instantiated DAG,
and the difficulty associated with the
corresponding reasoning state.

\begin{table}[t]
\centering
\small
\begin{tabular}{lc}
\hline
Parameter & Value \\
\hline
$\alpha / \beta / \gamma$ & $0.25 / 0.35 / 0.40$ \\
$\tau$ & $0.8$ \\
$w_{\min} / w_{\max}$ & $0.5 / 2.0$ \\
\hline
\end{tabular}
\caption{Hyperparameters used for DAG-guided weight assignment.}
\label{tab:weight_hyperparameters}
\end{table}

\paragraph{Weight normalization.}
We normalize the node scores using a
temperature-controlled softmax:
\begin{equation}
\tilde{w}_i =
\frac{\exp(z_i/\tau)}
{\sum_j \exp(z_j/\tau)},
\end{equation}
where $\tau$ controls the concentration of
the resulting weight distribution.
The normalized score is then mapped to a
bounded supervision range:
\begin{equation}
w_i =
w_{\min} +
(w_{\max}-w_{\min})\tilde{w}_i.
\end{equation}
Table~\ref{tab:weight_hyperparameters}
summarizes the fixed hyperparameters used for
weight assignment across all experiments.
This transformation preserves the relative
importance induced by the fused node scores
while bounding the contribution of individual
reasoning states.

\paragraph{Token-level application.}
Each node weight is assigned to the
target-token span corresponding to that
reasoning state.
Relation and answer spans use their designated
span weights, consistent with the training
objective in Section~\ref{Weighted DAG Supervision}.
Given target tokens $y_{1:T}$, the weighted
autoregressive objective is
\begin{equation}
\mathcal{L}_{\mathrm{DAG}}
=
\frac{
\sum_{t=1}^{T}
w_t
\mathcal{L}_{\mathrm{CE}}
(y_t,\hat{y}_t)
}{
\sum_{t=1}^{T} w_t
},
\end{equation}
where $w_t$ denotes the supervision weight
assigned to the span containing token $t$.
The structural matrices are used only to
construct the supervision weights and are
not provided as additional model inputs.

\subsection{Optimization and Fine-Tuning Settings}
\label{app:optimization}

We fine-tune Qwen3-VL-4B-Instruct using LoRA.
Table~\ref{tab:training_4b} summarizes the
task-specific optimization settings used for
open-ended and multiple-choice QA.
Gradient checkpointing is enabled during
fine-tuning.
Qwen3-VL-8B-Instruct and InternVL3.5-4B-Instruct
use the same task-specific training configuration.
All experiments were conducted on four NVIDIA RTX 3090 GPUs with 24 GB of memory each. The total computational budget, including preliminary experiments, was approximately 1,500 GPU-hours.
We implement our experiments using PyTorch 2.6.0,
Hugging Face Transformers 4.57.6, and PEFT 0.17.1.
\begin{table}[t]
\centering
\small
\begin{tabular}{lcc}
\hline
Setting & Open-ended & MCQ \\
\hline
LoRA rank / scaling & 32 / 64 & 32 / 64 \\
LoRA dropout & 0.05 & 0.05 \\
Learning rate & $5\times10^{-5}$ & $3\times10^{-5}$ \\
Schedule & Constant & Cosine \\
Warmup ratio & 0 & 0.03 \\
Weight decay & 0 & 0.01 \\
Global batch size & 16 & 16 \\
Epochs & 2 & 2 \\
Gradient clipping & 1.0 & 1.0 \\
\hline
\end{tabular}
\caption{Training settings for open-ended and
multiple-choice QA. The same task-specific
configuration is used across model backbones.}
\label{tab:training_4b}
\end{table}

\paragraph{Weighted supervision.}
We apply token-level cross-entropy to assistant
target tokens while masking input and padding
positions.
For VizDAGuide, reasoning spans receive the
supervision weights described in
Section~\ref{app:adaptive_weights}.
The Linear CoT baseline applies the same weights
to the corresponding semantic spans, whereas
the unweighted DAG assigns unit weights to its
target spans.
For weighted supervision, the loss is normalized
by the sum of the active token weights, following
the objective in Section~\ref{Weighted DAG Supervision}.

\section{Evaluation Metrics and Protocol}
\label{app:evaluation}

\subsection{Metric Definitions}
\label{app:metric_definitions}

Our evaluation criteria are aligned with the seven
reasoning states defined in
Section~\ref{abilities}.
Each state is associated with a corresponding
semantic check, while Answer Generation is evaluated
using the task-specific answer-level metric.
The main paper reports a subset of these criteria
for compactness.

\paragraph{Question Scoping.}
Question Scoping evaluates whether the generated
reasoning correctly identifies the task requirements
of the question, including the relevant entities,
comparison or relation, and the quantity or property
to be determined.
The check focuses on the semantic scope of the
question rather than the exact wording of the
generated field.

\paragraph{Evidence Grounding.}
Evidence Grounding evaluates whether the generated
reasoning identifies the evidence required to answer
the question.
For numerical evidence, predicted entities are
aligned with the corresponding reference entities,
and their required values are evaluated with a
5\% relative tolerance.
Conflicting values for the same entity fail the
check.
For qualitative MCQ instances, Grounding checks
whether the generated observation captures the
reference chart condition required to interpret
the question.

\paragraph{Visual-Risk Diagnosis.}
Visual-Risk Diagnosis evaluates whether the model
correctly identifies the visual condition relevant
to the required inference.
For misleading charts, the predicted diagnosis
must identify the misleading cue that can affect
the question.
For clean charts, the model should not introduce
a misleading condition that requires correction.
The metric evaluates the semantic diagnosis rather
than exact surface-form matching.

\paragraph{Correction Selection.}
Correction Selection evaluates whether the model
selects an appropriate response to the diagnosed
visual risk.
For misleading instances, the predicted correction
must be consistent with the reference correction
required for the question.
For clean instances, the model should determine
that no correction is needed.

\paragraph{Operation Execution.}
Operation Execution evaluates whether the generated
reasoning identifies and applies the operation
required by the question.
For open-ended QA, the evaluation considers the
operation type, the evidence used by the operation,
and the resulting computation.
When numerical execution is applicable, the
computed result is checked against the reference
with a 5\% relative tolerance.
For MCQ, the metric evaluates agreement with the
reference operation or reasoning category.

\paragraph{Decision Formation.}
Decision Formation evaluates whether the model
derives the correct task-level conclusion from the
preceding reasoning states before rendering the
final answer.
For numerical questions, this corresponds to the
derived numerical conclusion.
For qualitative or multiple-choice questions, it
corresponds to the semantic conclusion supported
by the grounded evidence and operation.

\paragraph{Answer Generation.}
Answer Generation is evaluated using the
task-specific answer-level metrics.
For open-ended QA, Strict Final-Answer Accuracy
(FAA) uses normalized text matching for textual
answers.
For numerical answers, a prediction is considered
correct when its absolute error is within
\begin{equation}
\max\left(0.001,\,
10^{-5}\max(|a|,1)\right),
\end{equation}
where $a$ denotes the reference answer.
Relaxed Accuracy (RA) follows the same procedure
but uses a numerical tolerance of
\begin{equation}
\max\left(0.001,\,0.05|a|\right).
\end{equation}
For MCQ, Accuracy measures whether the predicted
option matches the reference answer, while
Misleading-Choice Rate (MCR) measures how often
the model selects the annotated misleading option.
All answer-level metrics are computed on the
fixed answer-eligible sets described in
Section~\ref{app:data}.

\paragraph{Clean False Correction.}
For clean charts, Clean False Correction
(Clean FC) measures the proportion of instances
for which the model incorrectly determines that
a correction is required.
It complements Correction Selection by measuring
unnecessary intervention on instances where the
visual encoding does not require correction.
Clean FC is not applicable to the MCQ benchmark,
whose test set contains only misleading charts.

\paragraph{Sufficiency.}
For MCQ, Sufficiency evaluates whether the
evidence identified by the model is sufficient
to support its final decision.
It complements Decision Formation by measuring
whether the conclusion is supported by the
available evidence rather than final-option
correctness alone.

\paragraph{Process Accuracy.}
Process Accuracy measures whether all applicable
process-level checks defined above are jointly
correct for the same instance.
An instance is counted as correct only when every
applicable process check is correct; if any
applicable check fails, the instance is counted
as incorrect.
Answer Generation is evaluated separately using
the answer-level metrics described above.
\paragraph{Process--answer relation.}
For each process metric, we measure its association
with final-answer correctness using the Phi
coefficient ($\phi$), which is equivalent to the
Pearson correlation coefficient for two binary
variables.
For each instance, process correctness and
final-answer correctness are encoded as binary
indicators.
The correlation is computed separately for
Grounding, Operation, Correction, and Sufficiency
using only instances for which the corresponding
process metric is applicable; instances marked as
\texttt{NA} are excluded.
A positive $\phi$ indicates that process correctness
is associated with answer correctness, whereas a
value near zero indicates a weak association.
If either binary variable is constant over the
applicable instances, $\phi$ is undefined and
reported as \texttt{NA}.

\subsection{External-Model Evaluation Protocol}
\label{app:external_protocol}

External models are evaluated using a unified
structured-output protocol.
Each request contains the chart image and
question, together with the candidate answer
options for MCQ.
Reference answers, underlying data tables,
misleader labels, and process annotations are
not provided to the model.

The requested output contains fields corresponding
to evidence, operation, visual correction,
decision, and final answer.
This model-agnostic representation provides the
information required by our process-level metrics
without requiring external models to reproduce
the VizDAGuide serialization, node identifiers,
or internal correction tokens.

For correction evaluation, external models select
from a supplied set of correction descriptions
expressed in natural language, including a
no-correction option.
This is necessary because the external models are
not trained with the correction representation
used by VizDAGuide.
The supplied choices expose the semantics of the
candidate corrections rather than our internal
operator notation, allowing Correction to be
evaluated under a common output interface.

We use deterministic decoding where supported
and preserve the raw model responses for
evaluation.
If a final answer can be reliably recovered from
an otherwise malformed structured response, it
remains eligible for answer-level scoring.
Missing applicable process fields are counted as
incorrect for the corresponding process metric
rather than removed from its denominator.

Our trained models generate their native
structured outputs, whereas external models
follow the unified evaluation protocol described
above.
Format-specific parsers map both output formats
to the common fields required by the evaluation
criteria.
Because the prompting and output representations
are not identical, comparisons with external
models are intended to provide performance
context rather than a strictly protocol-matched
model ranking.

\subsection{External Evaluation Protocol Example}
\label{app:protocol_example}

We illustrate the external evaluation protocol
using the reversed-axis MCQ shown in
Figure~\ref{fig:example}.
External models receive the chart image,
question, and answer options without reference
answers, underlying tables, misleader labels,
or process annotations.

\paragraph{Example input.}
\begin{quote}
What is the trend in the number of museums
reported in the chart from 2006 to 2020?

A. Decreasing \quad B. Increasing

C. Fluctuating \quad D. Constant
\end{quote}

\paragraph{Protocol instructions (abbreviated).}
The following instructions summarize the
actual evaluation prompt.

\begin{quote}
Answer using only the supplied chart image,
question, and answer options.
Return a JSON object containing observations,
calculation, visual check, decision, and final
answer.

Record the evidence required by the question
without inventing unreadable precision.
Reference the recorded observations using
\texttt{v0}, \texttt{v1}, etc. in the calculation.

Select a correction from the supplied
natural-language descriptions, including a
no-correction option.
Do not use training-specific operators,
DAG nodes, or dependency edges.

For MCQ, the decision and final answer must
specify the same option letter.
\end{quote}

\paragraph{Illustrative JSON response.}
\begingroup
\footnotesize
\begin{verbatim}
{
  "observations": [
    {
      "series": "Museums",
      "category": "2006",
      "value": 55
    },
    {
      "series": "Museums",
      "category": "2020",
      "value": 5
    }
  ],
  "calculation": {
    "type": "trend",
    "expression": "v1 < v0",
    "result": true,
    "explanation": "The value decreases."
  },
  "visual_check": {
    "needed": true,
    "choice": "C01",
    "reason": "Reversed vertical axis."
  },
  "decision": "A",
  "final_answer": "A"
}
\end{verbatim}
\endgroup

The instructions above are abbreviated from
the actual evaluation prompt.
The JSON response illustrates the required
output format and is not an actual model
prediction.
In this example, \texttt{C01} denotes the
reversed-axis correction.
Correction descriptions are shuffled
deterministically by sample ID, so their
identifiers do not have fixed meanings across
instances.

For open-ended QA, the decision and final
answer contain a numerical or textual
response rather than a candidate option.

\section{Statistical Analysis}
\label{app:statistics}

\subsection{Paired Model Comparisons}
\label{app:model_significance}

To assess the main comparisons in
Table~\ref{tab:ablation}, we align predictions
by sample ID and use the same reference
annotations and eligibility masks for the
compared systems.
The analysis covers 5,961 answer-eligible
open-ended questions from 4,113 chart groups
and 880 answer-eligible MCQ questions from
836 chart groups.
Differences are reported in percentage points
as VizDAGuide minus the corresponding
comparator.

We estimate nominal 95\% percentile confidence
intervals using 10,000 paired chart-group
bootstrap replicates with seed 7.
Questions associated with the same chart are
resampled together.
Statistical significance is assessed using
10,000 two-sided chart-group sign-flip
permutations.
For $B=10{,}000$ permutations, the permutation
$p$-value is estimated as
\begin{equation}
p =
\frac{k+1}{B+1},
\end{equation}
where $k$ is the number of permuted differences
at least as extreme as the observed difference.
We additionally compute two-sided exact McNemar
tests on paired answer outcomes.
We report the chart-group permutation results
because multiple questions associated with the
same chart may share visual information.

\begin{table*}[t]
\centering
\small
\begin{tabular}{lllcc}
\hline
Model & Comparator & Metric &
$\Delta$ [95\% CI] & $p$ \\
\hline
\multicolumn{5}{c}{Open-ended QA} \\
\hline
Qwen3-VL-4B & Linear CoT & FAA
& +1.29 [0.45, 2.12] & 0.0020 \\
Qwen3-VL-4B & Linear CoT & RA
& +1.69 [0.82, 2.55] & 0.0001 \\
Qwen3-VL-8B & Linear CoT & FAA
& +2.43 [1.63, 3.23] & 0.0001 \\
Qwen3-VL-8B & Linear CoT & RA
& +2.42 [1.57, 3.25] & 0.0001 \\
InternVL3.5-4B & Linear CoT & FAA
& +1.14 [0.33, 1.92] & 0.0070 \\
InternVL3.5-4B & Linear CoT & RA
& +1.09 [0.25, 1.88] & 0.0092 \\
\hline
Qwen3-VL-4B & Base & FAA
& +22.56 [21.32, 23.84] & 0.0001 \\
Qwen3-VL-4B & Base & RA
& +24.95 [23.65, 26.27] & 0.0001 \\
Qwen3-VL-8B & Base & FAA
& +17.60 [16.40, 18.80] & 0.0001 \\
Qwen3-VL-8B & Base & RA
& +19.33 [18.05, 20.59] & 0.0001 \\
InternVL3.5-4B & Base & FAA
& +13.05 [11.81, 14.27] & 0.0001 \\
InternVL3.5-4B & Base & RA
& +15.47 [14.17, 16.73] & 0.0001 \\
\hline
\multicolumn{5}{c}{Multiple-choice QA} \\
\hline
Qwen3-VL-4B & Linear CoT & Acc.
& +2.61 [0.23, 5.07] & 0.0471 \\
Qwen3-VL-4B & U-DAG & Acc.
& +3.86 [1.81, 5.98] & 0.0005 \\
Qwen3-VL-8B & Linear CoT & Acc.
& +2.27 [-0.11, 4.68] & 0.0679 \\
InternVL3.5-4B & Linear CoT & Acc.
& +2.16 [-0.23, 4.55] & 0.0975 \\
\hline
Qwen3-VL-4B & Base & Acc.
& +11.82 [8.39, 15.25] & 0.0001 \\
Qwen3-VL-8B & Base & Acc.
& +20.11 [16.17, 24.22] & 0.0001 \\
InternVL3.5-4B & Base & Acc.
& +20.34 [16.53, 24.04] & 0.0001 \\
\hline
\end{tabular}
\caption{Paired significance analysis for the main
comparisons in Table~\ref{tab:ablation}.
Differences are reported as VizDAGuide minus the
corresponding comparator in percentage points.
Confidence intervals use paired chart-group
bootstrap, and $p$-values use two-sided
chart-group permutation tests.
The reported $p$-values are not adjusted for
multiple comparisons.
A value of 0.0001 is the minimum attainable
permutation estimate rounded to four decimal
places.}
\label{tab:model_significance}
\end{table*}

\paragraph{Results.}
On open-ended QA, VizDAGuide improves both FAA
and RA over the matched Linear CoT baseline
for Qwen3-VL-4B, Qwen3-VL-8B, and
InternVL3.5-4B, with all confidence intervals
excluding zero.
On MCQ, the Qwen3-VL-4B improvement over
Linear CoT reaches the unadjusted 0.05 threshold,
and its improvement over U-DAG also passes this
threshold.
The corresponding Qwen3-VL-8B and
InternVL3.5-4B differences over Linear CoT
are positive but do not pass the unadjusted
0.05 threshold.
All reported comparisons against the Base
models pass the unadjusted threshold.
The Linear CoT and U-DAG comparisons directly
evaluate the supervision variants, whereas
comparisons with Base characterize the overall
gain of the trained system.

\subsection{Fine-grained Comparisons and Holm Correction}
\label{app:fine_grained_statistics}

For the fine-grained Qwen3-VL-4B comparison
between VizDAGuide and Base in
Section~\ref{sec:fine_grained}, predictions
are paired by sample ID.
We use two-sided exact McNemar tests for
category-level answer correctness and apply
Holm correction separately within each
dataset.

Within each dataset, the correction jointly
covers all question-type, chart-type, and
misleader-type comparisons shown in
Figure~\ref{fig:comparison}.
This gives 20 comparisons for open-ended QA
and 29 comparisons for MCQ.

For ordered raw $p$-values
$p_{(1)}\leq\cdots\leq p_{(m)}$, the
Holm-adjusted values are computed as
\begin{equation}
p^{\mathrm{Holm}}_{(k)}
=
\min\left(
1,
\max_{j\leq k}
\left[(m-j+1)p_{(j)}\right]
\right).
\end{equation}
We use an adjusted significance threshold of
$0.05$.

\paragraph{Results.}
For open-ended QA, the improvements over Base
remain significant after Holm correction for
all five question types and all three chart
types.
Among the 12 misleading-condition categories,
11 remain significant after correction, while
Dual-axis Misuse does not pass the corrected
threshold.

For MCQ, the improvement for
Find Correlations/Trends remains significant
after Holm correction.
Among chart types, Area Chart and Line Chart
remain significant.
Among misleading conditions,
Unconventional Scale Direction and
Inappropriate Aggregation remain significant.
The remaining category-level differences do not
pass the Holm-adjusted threshold.

\subsection{Process--Answer Association}
\label{app:process_answer}

We use the Phi coefficient ($\phi$) to measure
the association between binary process
correctness and final-answer correctness.
For two binary variables, the Phi coefficient
is equivalent to the Pearson correlation
coefficient:
\begin{equation}
\phi =
\frac{
n_{11}n_{00}-n_{10}n_{01}
}{
\sqrt{
(n_{11}+n_{10})
(n_{01}+n_{00})
(n_{11}+n_{01})
(n_{10}+n_{00})
}
},
\end{equation}
where $n_{uv}$ denotes the number of instances
with process correctness $u$ and answer
correctness $v$.

For each instance, process correctness and
final-answer correctness are encoded as binary
indicators.
The coefficient is computed separately for
each applicable process metric, including
Grounding, Operation, Correction, and
Sufficiency where applicable.
Only instances for which the corresponding
process reference is available are included;
instances marked as \texttt{NA} are excluded.
If either binary variable has zero variance,
$\phi$ is undefined and is reported as
\texttt{NA}.

A positive $\phi$ indicates that process
correctness is associated with answer
correctness, whereas a value near zero
indicates a weak association.
The coefficient measures statistical
association rather than causal contribution
to final-answer accuracy.

\subsection{Single-stage Bottleneck Analysis}
\label{app:bottleneck}

To examine the relation between individual
process failures and final-answer errors, we
identify isolated failures for each
component-level process metric.
An isolated failure is an instance in which
the corresponding process check is incorrect
while all other applicable component-level
process checks are correct.

For each process component, we report the
final-answer error rate among these isolated
failures.
This quantity corresponds to the
\textit{Answer Error} measure reported in the
main-text process--answer analysis.
We additionally retain the number of applicable
instances when interpreting these conditional
error rates.

Process Accuracy is not included as a separate
conditioning variable because it is an aggregate
of the component-level process checks rather
than an independent reasoning component.

\section{Qualitative Analysis}
\label{app:qualitative}

We present qualitative comparisons between
VizDAGuide and Linear CoT across different
misleading chart conditions.
These examples illustrate how the two methods
differ in evidence grounding, visual correction,
and operation execution.

\begin{figure}[t]
    \centering
    \includegraphics[width=\columnwidth]{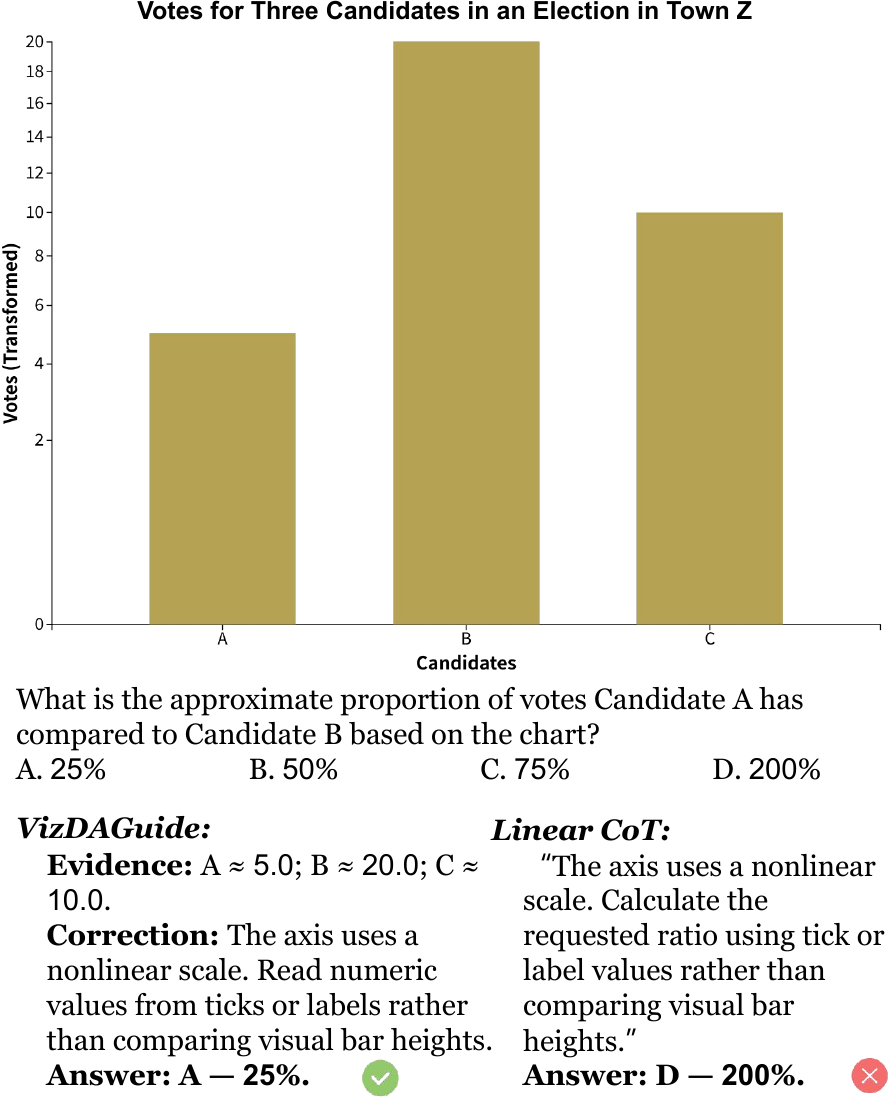}
    \caption{Qualitative comparison on a chart with a nonlinear scale.
    Both methods identify the misleading scale, but VizDAGuide
    correctly computes the vote ratio (25\%), whereas Linear CoT
    returns an incorrect answer (200\%).}
    \label{fig:appendix_nonlinear_votes}
\end{figure}

\paragraph{Nonlinear scale.}
As shown in Figure~\ref{fig:appendix_nonlinear_votes},
both methods recognize the nonlinear axis.
However, Linear CoT fails to obtain the
correct ratio, while VizDAGuide grounds
the numerical values and produces the
correct answer.
This example highlights the importance of
carrying grounded evidence into the
required operation.

\begin{figure}[t]
    \centering
    \includegraphics[width=\columnwidth]{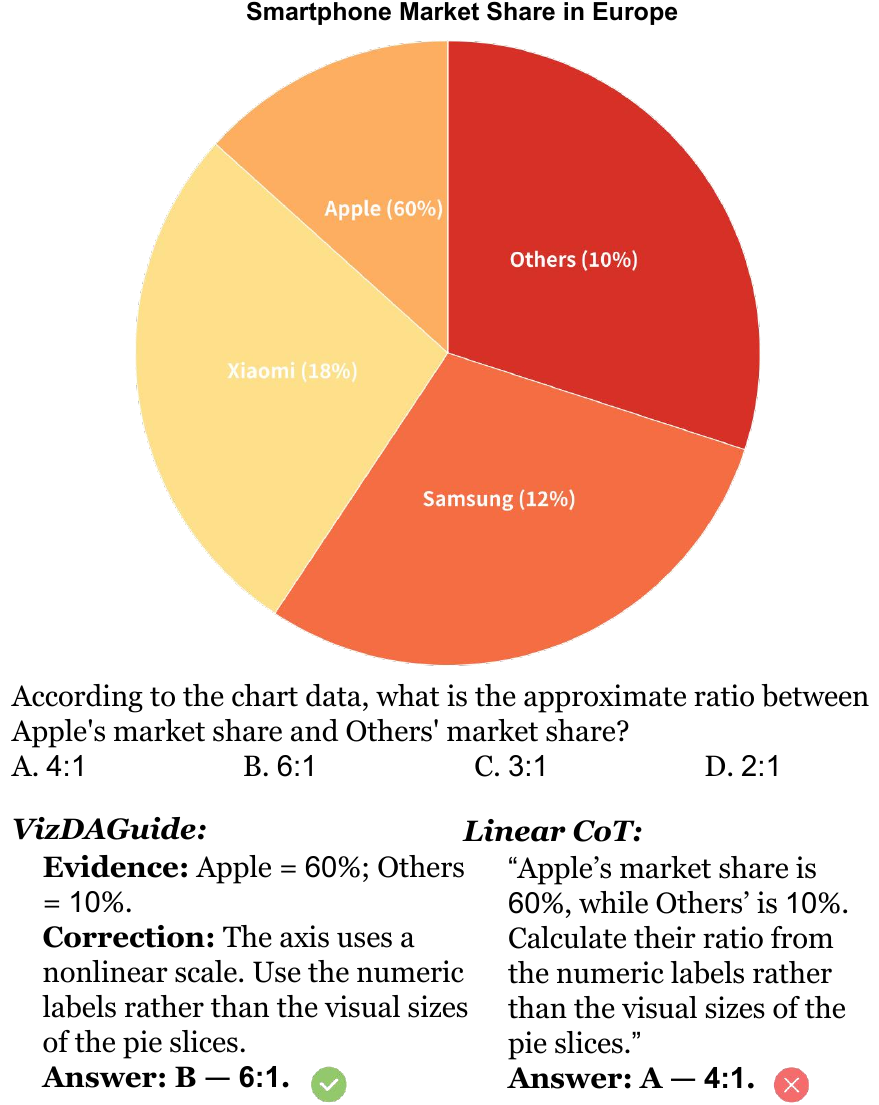}
    \caption{Qualitative comparison on a pie chart with misleading slice proportions.
    Both methods identify the labeled market shares, but VizDAGuide
    returns the correct ratio (6:1), whereas Linear CoT
    selects an incorrect answer (4:1).}
    \label{fig:appendix_distorted_pie}
\end{figure}

\paragraph{Distorted pie slices.}
As shown in Figure~\ref{fig:appendix_distorted_pie},
both methods identify Apple's market share as
60\% and Others' as 10\%.
However, Linear CoT selects an incorrect ratio
despite identifying the relevant values.
VizDAGuide uses the grounded values to obtain
the correct ratio of 6:1, illustrating better
consistency between evidence and operation.

\begin{figure}[t]
    \centering
    \includegraphics[width=\columnwidth]{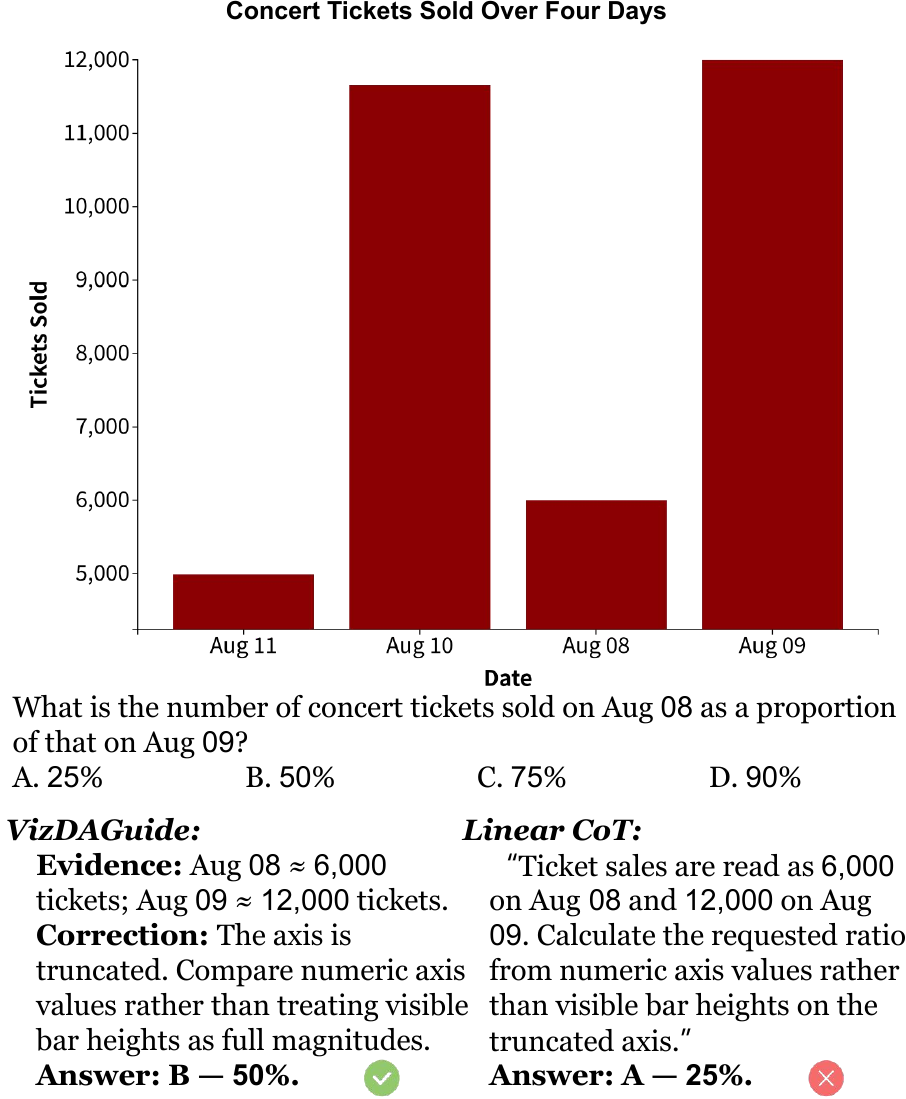}
    \caption{Qualitative comparison on a bar chart with a truncated axis.
    Both methods identify the truncated scale and the relevant values,
    but VizDAGuide returns the correct ratio (50\%), whereas
    Linear CoT selects an incorrect answer (25\%).}
    \label{fig:appendix_truncated_tickets}
\end{figure}

\paragraph{Truncated axis.}
As shown in Figure~\ref{fig:appendix_truncated_tickets},
both methods recognize the truncated axis and
identify the ticket sales as 6,000 and 12,000.
However, Linear CoT returns an incorrect
proportion despite stating the relevant values.
VizDAGuide produces the correct ratio of 50\%,
illustrating better consistency between
Evidence Grounding and the final answer.

\begin{figure}[t]
    \centering
    \includegraphics[width=\columnwidth]{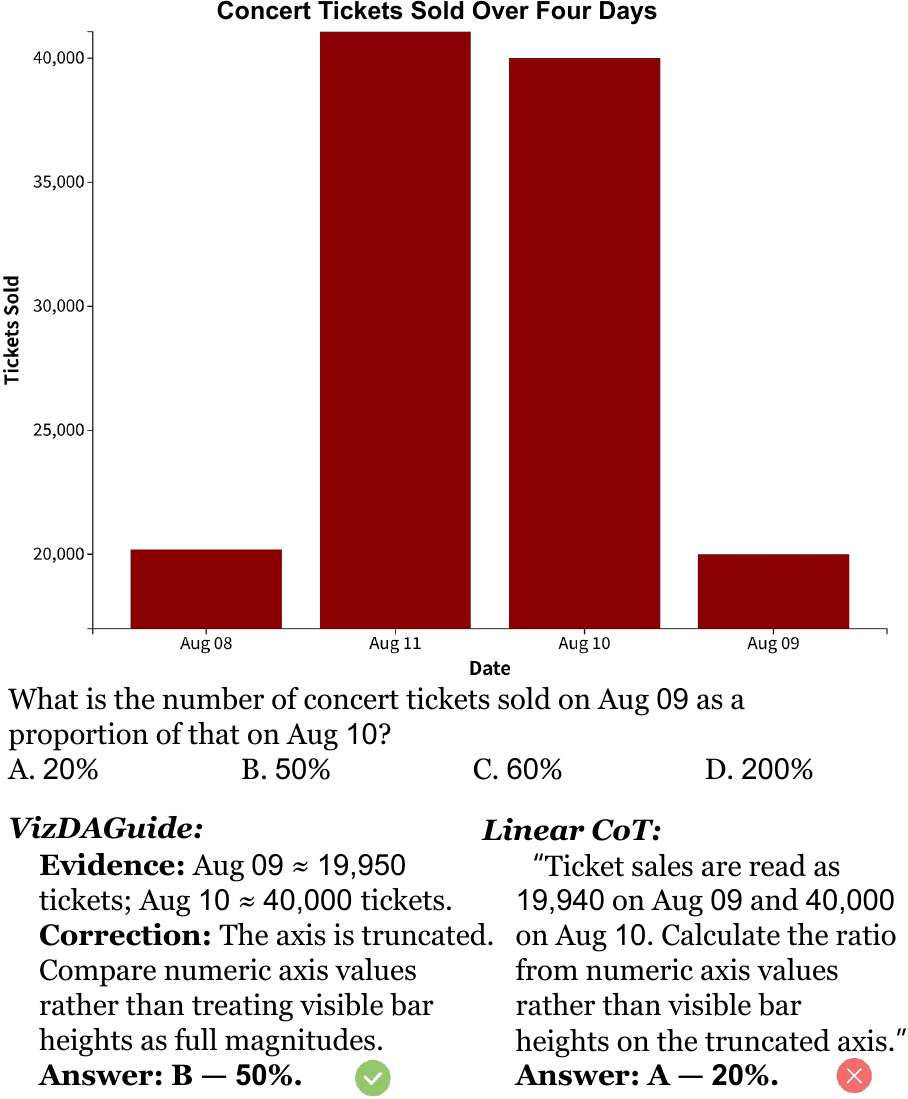}
    \caption{Qualitative comparison on a truncated-axis chart.
    VizDAGuide computes the correct ticket-sales ratio
    (50\%) from the numerical values, whereas Linear CoT
    selects an incorrect answer (20\%).}
    \label{fig:appendix_truncated_ratio}
\end{figure}

\paragraph{Evidence and operation consistency.}
Figures~\ref{fig:appendix_nonlinear_votes}--
\ref{fig:appendix_truncated_ratio} illustrate
cases involving numerical ratios under misleading
visual encodings.
In these examples, Linear CoT identifies the
relevant values or correction strategy but
produces an incorrect ratio.
VizDAGuide instead derives the correct answers
from the grounded numerical evidence.
These cases illustrate differences in how
evidence is carried through Operation Execution.

\begin{figure}[t]
    \centering
    \includegraphics[width=\columnwidth]{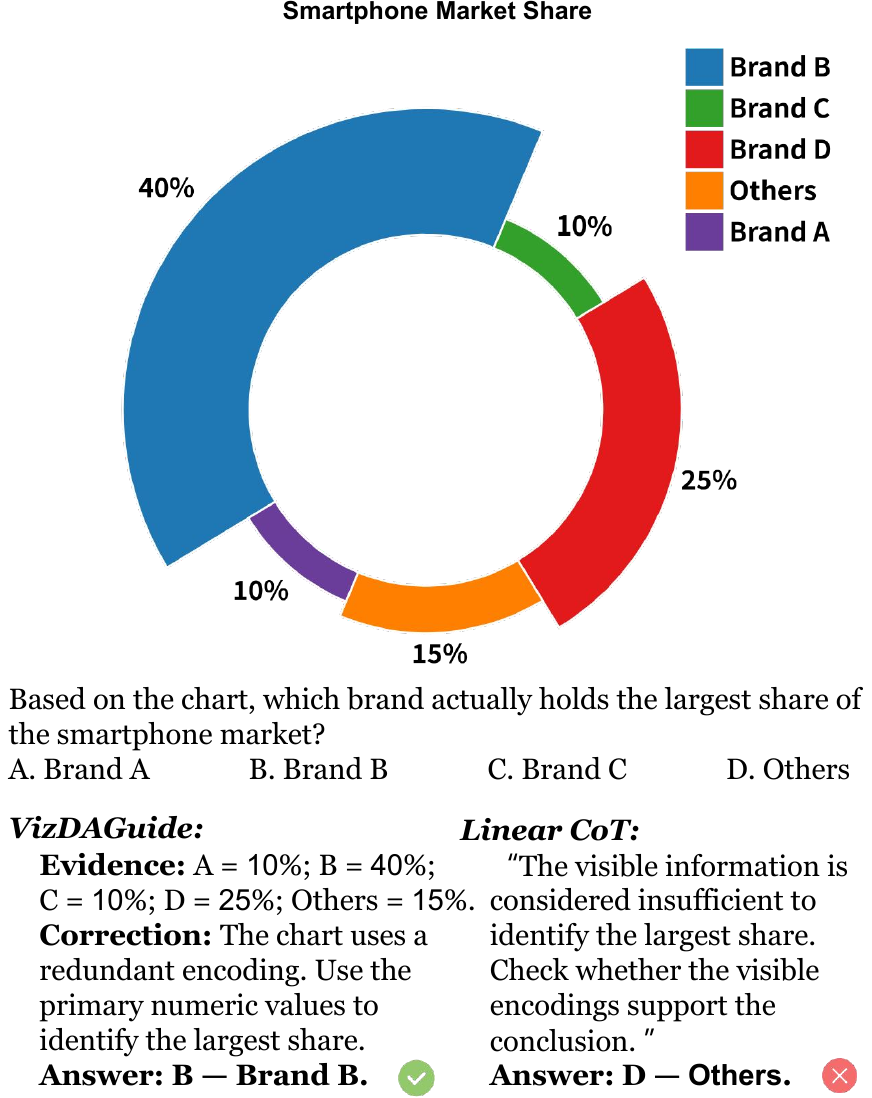}
    \caption{Qualitative comparison on a chart with redundant
    visual encoding. VizDAGuide identifies Brand B as having
    the largest market share (40\%), whereas Linear CoT
    incorrectly selects Others.}
    \label{fig:appendix_dual_encoding}
\end{figure}

\begin{figure}[t]
    \centering
    \includegraphics[width=\columnwidth]{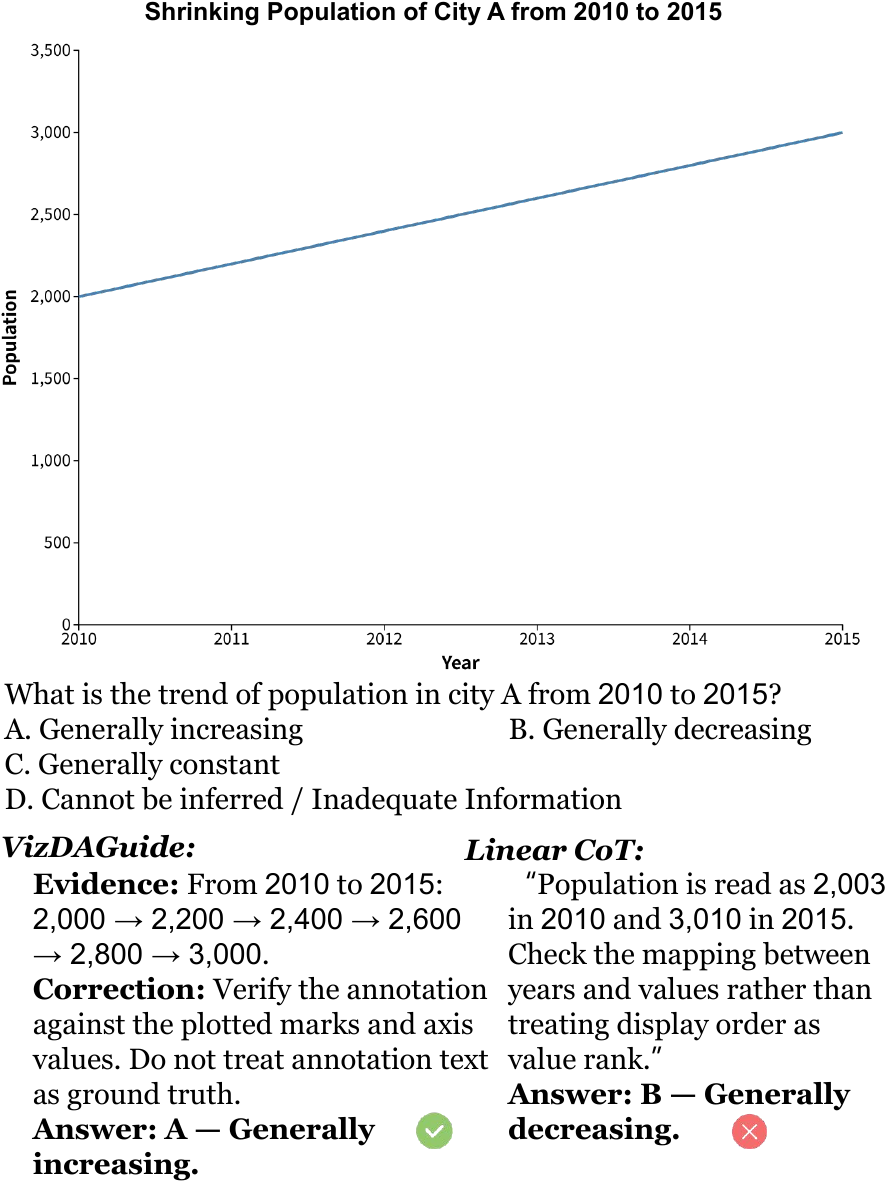}
    \caption{Qualitative comparison on a chart with a
    misleading title. Despite the title suggesting a
    population decline, the plotted values increase.
    VizDAGuide identifies the correct trend, whereas
    Linear CoT predicts a decrease.}
    \label{fig:appendix_misleading_title}
\end{figure}

\begin{figure}[t]
    \centering
    \includegraphics[width=\columnwidth]{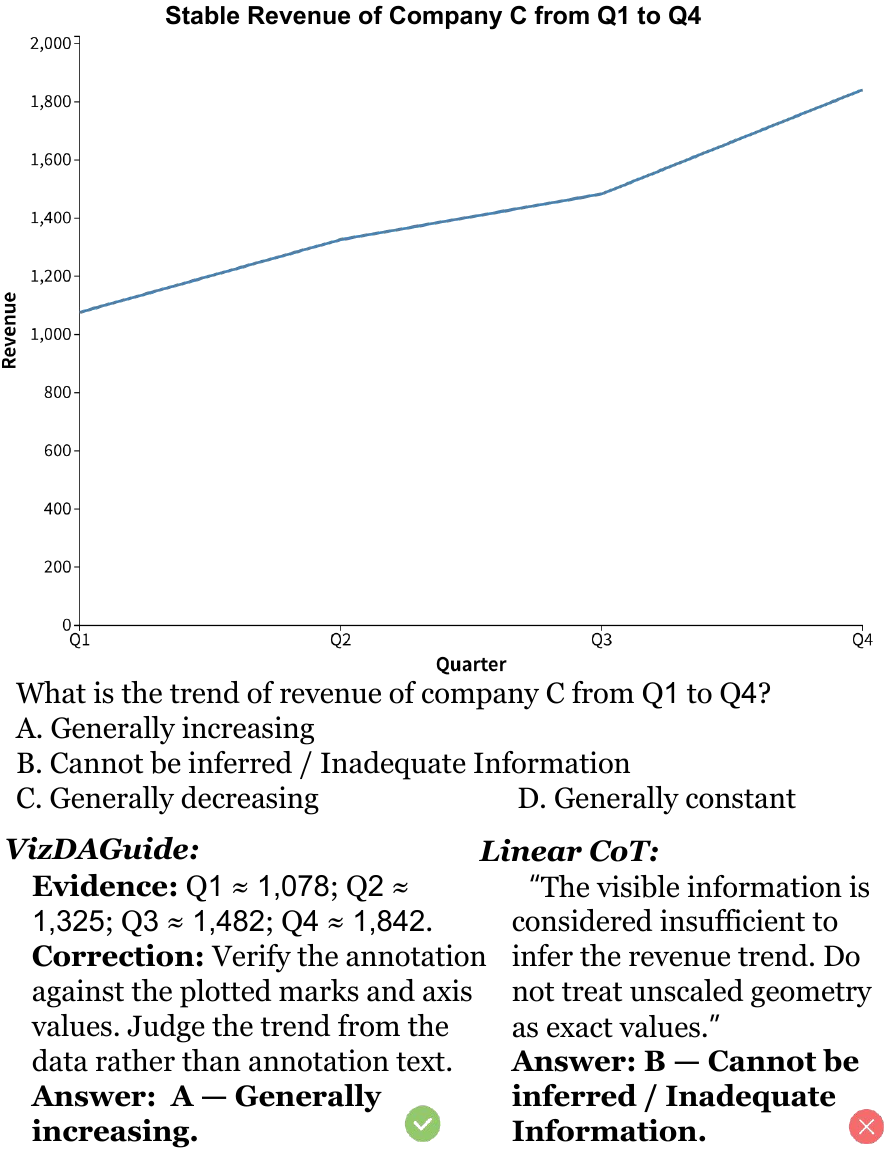}
    \caption{Qualitative comparison on a revenue chart
    with a misleading title. VizDAGuide identifies the
    increasing trend from the plotted values, whereas
    Linear CoT incorrectly concludes that the available
    information is insufficient.}
    \label{fig:appendix_revenue_trend}
\end{figure}

\begin{figure}[t]
    \centering
    \includegraphics[width=\columnwidth]{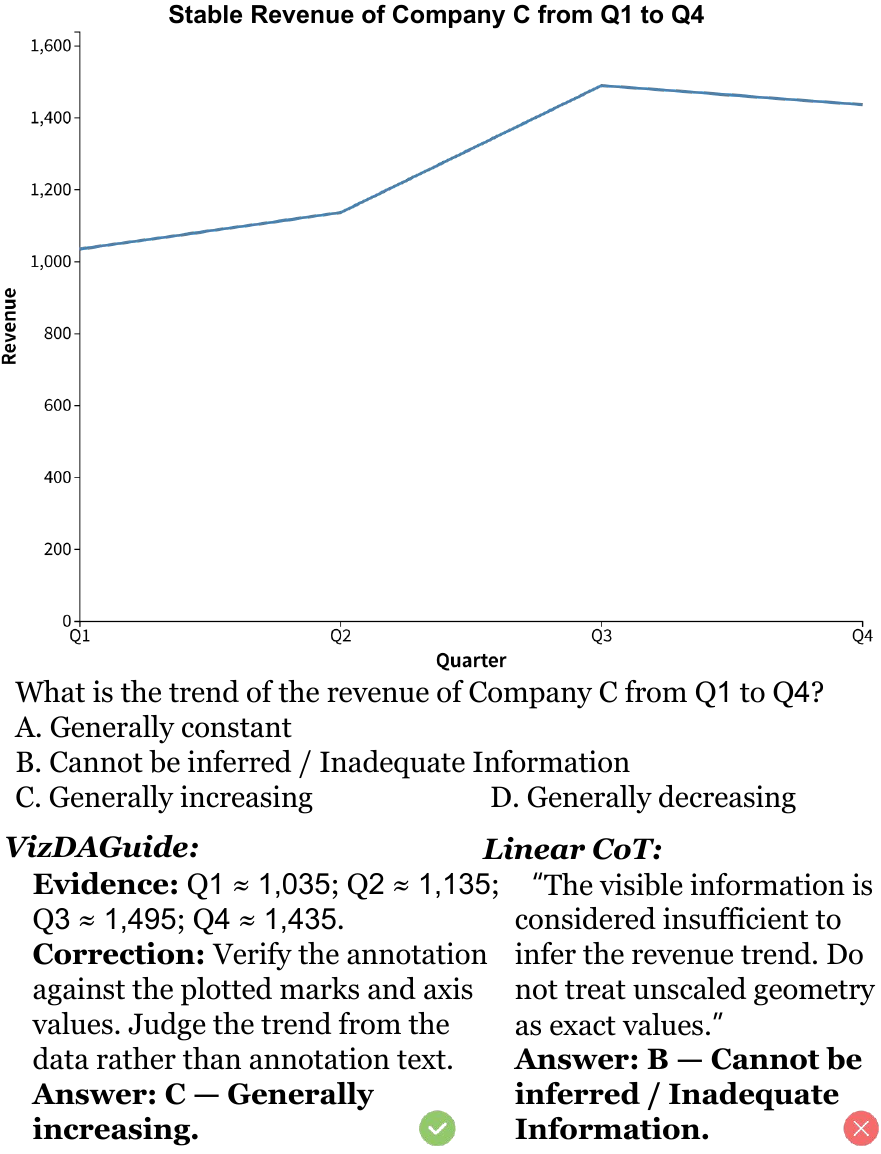}
    \caption{Qualitative comparison of overall trend
    interpretation. VizDAGuide identifies a generally
    increasing trend despite a decrease in the final
    interval, whereas Linear CoT selects the
    insufficient-information option.}
    \label{fig:appendix_overall_trend}
\end{figure}

\paragraph{Visual encoding and misleading titles.}
Figures~\ref{fig:appendix_dual_encoding}--
\ref{fig:appendix_overall_trend} present examples
involving redundant visual encoding and
misleading chart titles.
VizDAGuide identifies the required categories
or trends from the chart data, while Linear CoT
either selects an incorrect option or treats
the available information as insufficient.
In particular, the trend examples show that
chart titles should not override the numerical
patterns represented by the plotted marks.

\begin{figure}[t]
    \centering
    \includegraphics[width=\columnwidth]{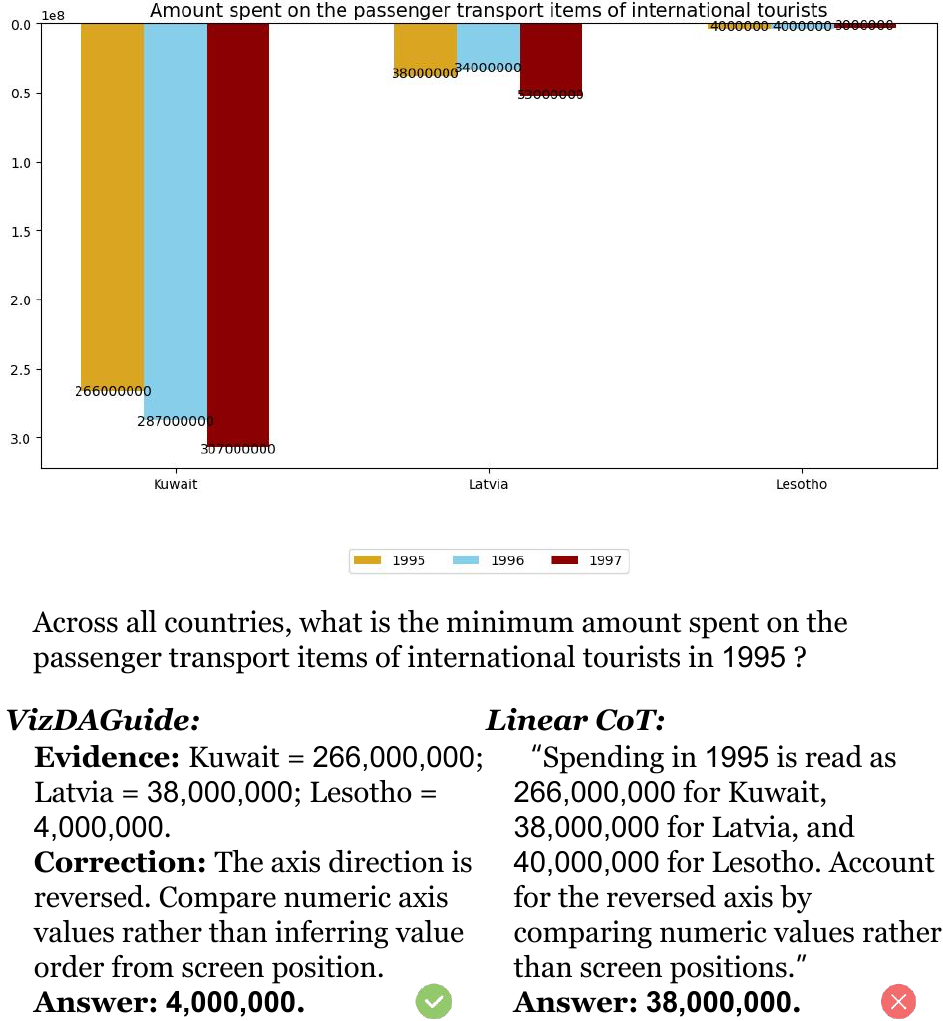}
    \caption{Open-ended QA comparison on a chart with
    an inverted axis. VizDAGuide identifies the correct
    minimum value (4,000,000), whereas Linear CoT
    returns 38,000,000.}
    \label{fig:appendix_oeq_inverted_minimum}
\end{figure}

\begin{figure}[t]
    \centering
    \includegraphics[width=\columnwidth]{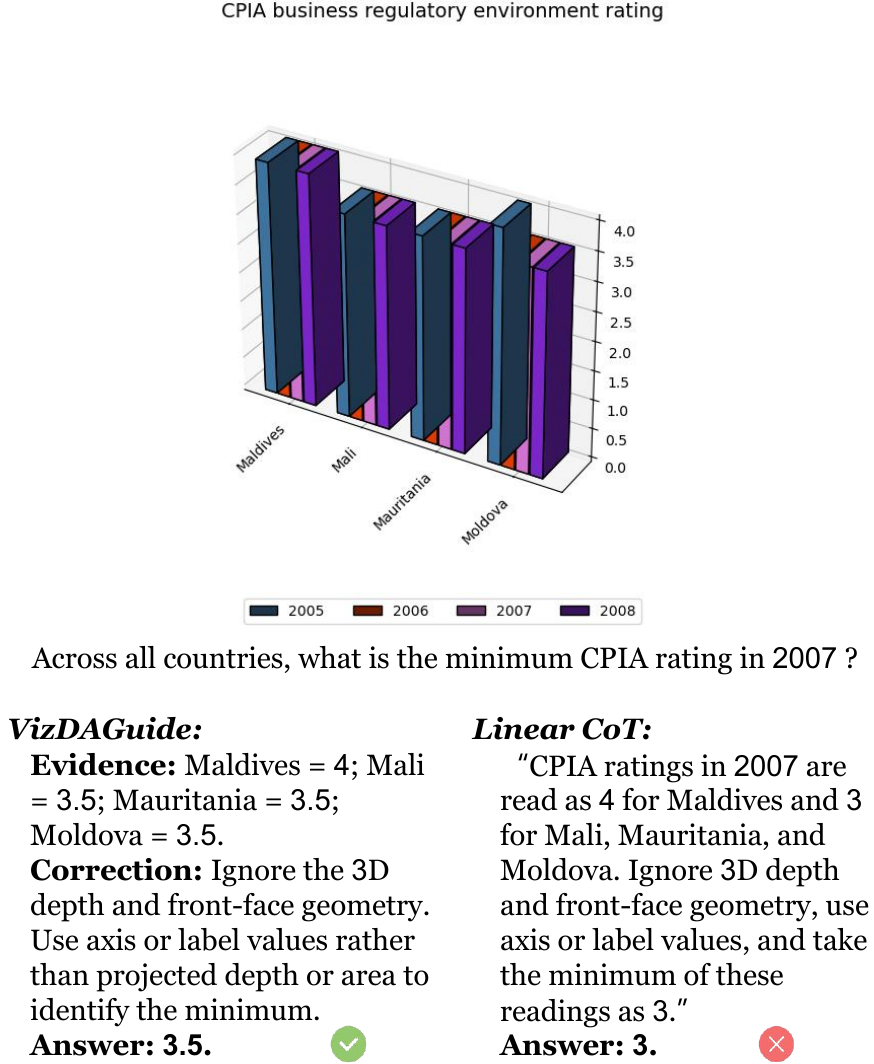}
    \caption{Open-ended QA comparison on a 3D bar chart.
    VizDAGuide identifies the correct minimum CPIA
    rating (3.5), whereas Linear CoT returns 3 after
    incorrectly reading the chart values.}
    \label{fig:appendix_oeq_3d_minimum}
\end{figure}

\begin{figure}[t]
    \centering
    \includegraphics[width=\columnwidth]{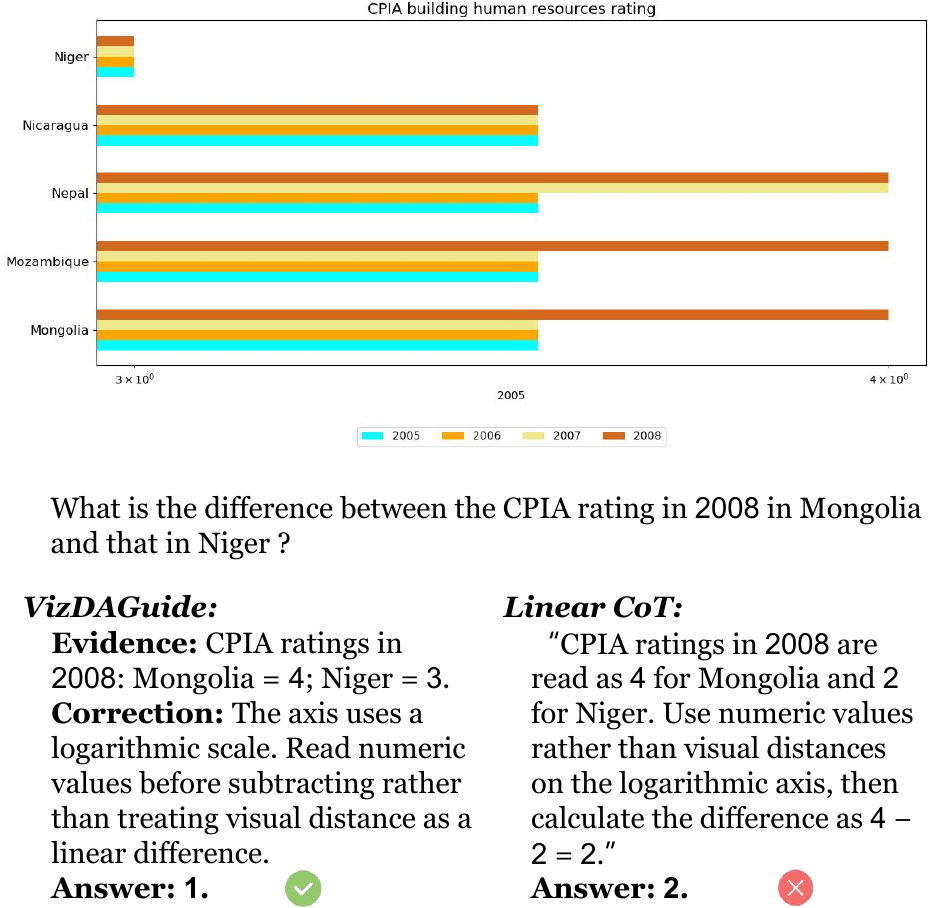}
    \caption{Open-ended QA comparison on a logarithmic-axis
    chart. VizDAGuide grounds the relevant values and
    computes the correct difference (1), whereas
    Linear CoT returns 2.}
    \label{fig:appendix_oeq_log_difference}
\end{figure}

\begin{figure}[t]
    \centering
    \includegraphics[width=\columnwidth]{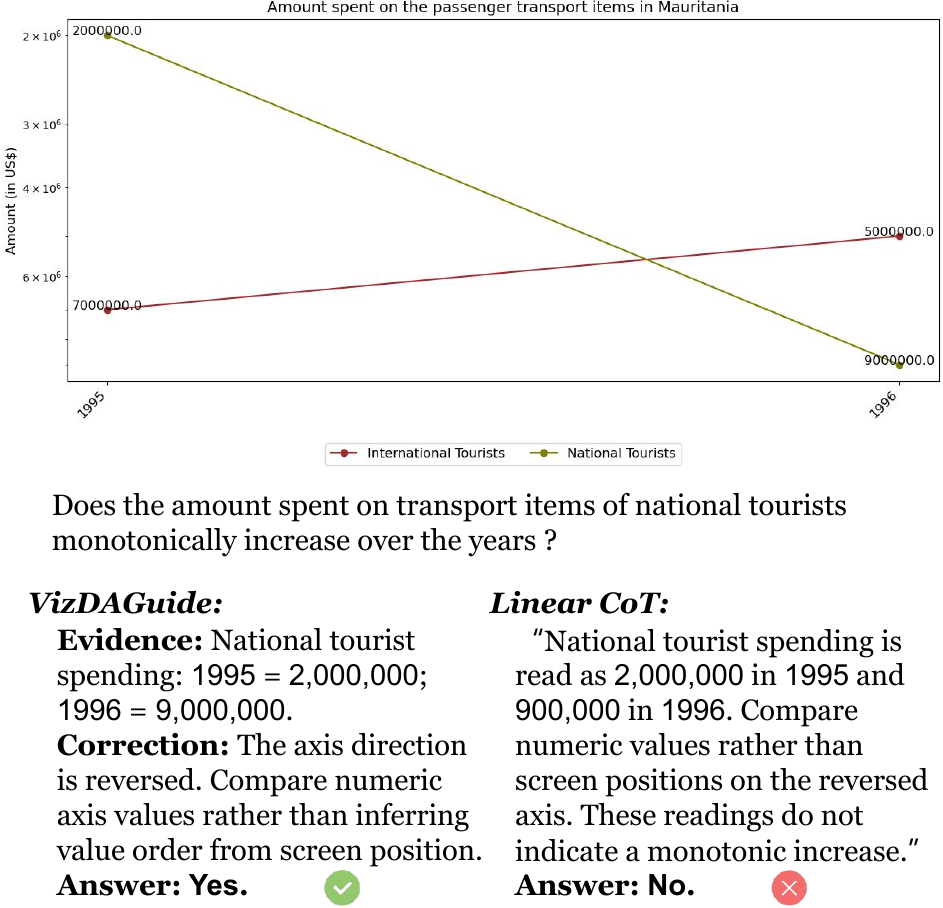}
    \caption{Open-ended QA comparison on an inverted-axis
    line chart. VizDAGuide correctly identifies the
    increasing numerical trend, whereas Linear CoT
    predicts the opposite trend.}
    \label{fig:appendix_oeq_inverted_trend}
\end{figure}

\paragraph{Numerical grounding in open-ended QA.}
Figures~\ref{fig:appendix_oeq_inverted_minimum}--
\ref{fig:appendix_oeq_inverted_trend} illustrate
open-ended questions involving inverted axes,
3D effects, and logarithmic scales.
VizDAGuide obtains the correct answers by
reading the relevant numerical values and
applying the required comparison or operation.
Linear CoT exhibits errors in value extraction
or numerical interpretation despite recognizing
some of the misleading visual conditions.
These examples complement the process-level
analysis by illustrating the role of Evidence
Grounding and Operation Execution in
final-answer correctness.


\end{document}